%% file: main.tex
\documentclass[11pt]{article}
\usepackage[final]{acl}
\usepackage{tipa}
\usepackage[utf8]{inputenc}
\usepackage[T1]{fontenc}
\usepackage[most]{tcolorbox}
\usepackage{times}       %
\usepackage{latexsym}
\usepackage{graphicx}
\usepackage{booktabs}
\usepackage{multirow}
\usepackage{makecell}
\usepackage{longtable}
\usepackage{microtype}

\usepackage{amsmath,amssymb}

\usepackage{algorithm}
\usepackage{algorithmic}
\usepackage{listings}

\title{The Sound of Syntax: Finetuning and Comprehensive Evaluation of Language Models for Speech Pathology}
\author{Fagun Patel$^{1*}$, Duc Q. Nguyen$^{2*}$, Sang T. Truong$^{1*}$,\\
\textbf{Jody Vaynshtok$^3$, Sanmi Koyejo$^1$, Nick Haber$^1$}\\
$^1$Stanford University $^2$National University of Singapore $^3$Sound Speech and Hearing Clinic\\$*$ Co-first Authors
}

\begin{document}
\maketitle
\input{sections/0abstract}
\input{sections/1intro}
\input{sections/2relatedwork}
\input{sections/3method}
\input{sections/4result}
\input{sections/5discussion}

\input{sections/7ack}

\bibliography{references}
\appendix
\onecolumn
\input{sections/6appendix}
\end{document}

%% file: sections/0abstract.tex
\begin{abstract}
    According to the U.S. National Institutes of Health, more than 3.4 million children experience speech disorders that require clinical intervention. The number of speech-language pathologists (SLPs) is roughly 20 times fewer than the number of affected children, highlighting a significant gap in children's care and a pressing need for technological support that improves the productivity of SLPs. State-of-the-art multimodal language models (MLMs) show promise for supporting SLPs, but their use remains underexplored largely due to a limited understanding of their performance in high-stakes clinical settings. To address this gap, we collaborate with domain experts to develop a taxonomy of real-world use cases of MLMs in speech-language pathologies. Building on this taxonomy, we introduce the first comprehensive benchmark for evaluating MLM across five core use cases, each containing 1,000 manually annotated data points. This benchmark includes robustness and sensitivity tests under various settings, including background noise, speaker gender, and accent. Our evaluation of 15 state-of-the-art MLMs reveals that no single model consistently outperforms others across all tasks. Notably, we find systematic disparities, with models performing better on male speakers, and observe that chain-of-thought prompting can degrade performance on classification tasks with large label spaces and narrow decision boundaries. Furthermore, we study fine-tuning MLMs on domain-specific data, achieving improvements of over 10\% compared to base models. These findings highlight both the potential and limitations of current MLMs for speech-language pathology applications, underscoring the need for further research and targeted development\footnote{To support continued progress, we publicly release our datasets, fine-tuned models, and benchmarking framework. Code: \url{https://github.com/AutonomousAgentsLab/slp_benchmark}, Dataset: \url{https://huggingface.co/datasets/SAA-Lab/SLPHelm}.}.
\end{abstract}

%% file: sections/1intro.tex
\section{Introduction}
Speech and language pathologies (SLP) in children can significantly impact communication, academic development, and long-term social outcomes~\citep{Hitchcock2015,Foster2023}. Early detection and intervention by speech-language pathologists are critical to mitigating these adverse effects~\citep{CDC2024}. Unfortunately, the availability of qualified clinicians is characterized by an uneven distribution across geographic and socioeconomic contexts, with only one expert for every 20 affected children, resulting in significant disparities in access to care and leading to ``missing intervention'' for many children who could benefit from timely support~\citep{usnih2025quick}. This gap highlights an urgent need for scalable and supportive technological solutions to augment clinicians' capabilities and expand the reach of vital interventions. 

The shortage of qualified clinicians has led to significant gaps in diagnostic capacity, particularly in domains that require specialized expertise. Recent advancements in multimodal large language models (MLMs) present a promising opportunity to partially automate or augment diagnostic workflows~\citep{Jessica2025,bhattacharya2024aphasia,nagpal2025speech,wagner24b_interspeech}. Multimodal LLMs, such as GPT-4 and Gemini, exhibit state-of-the-art capabilities in speech processing and contextual reasoning. Effective integration of LLMs into clinical SLP workflow requires rigorous evaluation to establish their clinical validity~\citep{Claire2025An}. This process relies on comprehensive and high-quality datasets that capture the variability of pediatric speech, especially disordered forms, and are annotated with clinical features. Currently, the evaluation is hindered by two key challenges: the scarcity of well-curated pediatric speech corpora and the lack of comprehensive evaluation frameworks for analyzing children’s speech.

In this study, we present a comprehensive evaluation of MLMs in SLP. We develop a systematic procedure for annotating symptoms and disorders in child speech, creating resources suitable for SLP-focused model evaluation and fine-tuning. The benchmark assesses models across five clinical scenarios covering a spectrum of tasks from foundational disorder detection to more granular symptoms, including Disorder Diagnosis, Transcription-Based Diagnosis, Transcription, Disorder Type Classification, and Symptom Classification. We systematically evaluate the capabilities and limitations of existing MLMs in SLP-relevant contexts and explore approaches for domain-specific adaptation. Our analysis reveals substantial performance gaps, with macro-F1 scores frequently falling below clinically acceptable thresholds, particularly on fine-grained tasks such as Disorder Type and Symptom Classification. To mitigate these shortcomings, we develop and assess fine-tuned MLMs, demonstrating improvements and advancing the state of the art on these specialized clinical tasks. Our contributions are summarized as follows.
\begin{itemize}%
    \item We release four curated pediatric speech datasets comprising approximately 30,000 speech samples across English and French, encompassing both typical and disordered speech. These datasets provide a publicly available, high-quality resource to support reproducible benchmarking in SLP.
    \item We introduce the first comprehensive evaluation framework for SLP, encompassing five essential clinical tasks. This framework enables consistent, task-aligned evaluation and facilitates direct comparison of speech LLM performance under a standardized protocol.
    \item We introduce fine-tuned speech LLMs that achieve state-of-the-art performance on disorder diagnosis and transcription tasks, illustrating the efficacy of domain-specific adaptation in enhancing diagnostic and transcriptional capabilities.
    \item We conduct extensive fine-grained analyses on model performance across various conditions, including demographic factors (gender, age), languages, and reasoning paradigms (e.g., Chain-of-Thought), offering deeper insights about model robustness and biases.
\end{itemize}

%% file: sections/2relatedwork.tex
\section{Related Works}
\paragraph{Comprehensive AI Benchmarking}
Comprehensive benchmarking has been instrumental in advancing speech-health research. The ADReSS Challenge~\cite{luz2020adress} established a balanced benchmark for Alzheimer’s detection from spontaneous speech, standardizing evaluation via F1 and MMSE-regression metrics. Similarly, the Children’s ASR Benchmark~\cite{fan2024childasr} introduced standardized splits and Whisper/Wav2Vec baselines for speech recognition in children aged 6–14, highlighting age-specific acoustic challenges. Nonetheless, systematic benchmarking of MLMs in clinical contexts remains limited.

\paragraph{Finetuning Audio Models}
Recent approaches have shown finetuning led to improvement in LLM audio understanding performance, especially for low-resource languages~\cite{pillai2024multistagefinetuningstrategiesautomatic}. Models finetuned via approaches like Continued Pretraining~\cite{ke2023continualpretraininglanguagemodels} and Reasoning Preference Optimization (RPO)~\cite{pang2024iterativereasoningpreferenceoptimization} have showcased improved reliability, factual accuracy, and cross-language generalization in clinical use cases~\cite{kawakami2025stabilizingreasoningmedicalllms}. Additionally, finetuned LLMs have been shown to outperform classical DNN approaches across clinical use cases such as murmur detection~\cite{florea2025exploringfinetunedaudiollmheart}. Furthermore, research has shown that while fine-tuning on specialized medical data enhances domain-specific knowledge, it can negatively impact a model's long-context understanding~\cite{yang2024finetuningmedicallanguagemodels}, highlighting the need for a balanced data composition during the fine-tuning process.

%% file: sections/3method.tex
\section{Method}
\subsection{Datasets, Annotation, and Core Scenarios}

\begin{figure*}[htbp]
  \centering
  \includegraphics[width=1\textwidth]{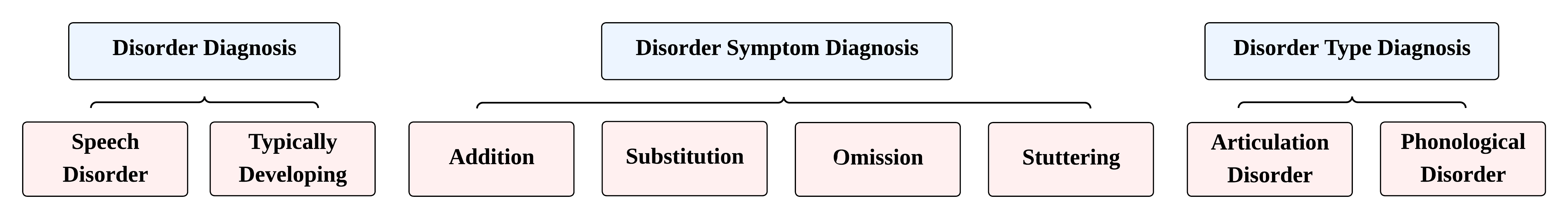} 
  \caption{Taxonomy of classification tasks in SLPHelm. The benchmark includes three core diagnostic tasks: (i) disorder diagnosis, (ii) disorder symptom classification, and (iii) disorder type classification.}
  \label{fig:taxonomy}
\end{figure*}

We use four datasets including Ultrasuite~\cite{eshky2018ultrasuite}, ENNI~\cite{schneider2006storytelling}, LeNormand~\cite{lenormand1997frenchmorph}, and Percept-GFTA~\cite{benway2022perceptr}. These datasets are publicly available, having been collected with informed consent and subsequently anonymized by their creators to protect participant privacy. The datasets encompass a range of child speech samples, both typical and disordered, and serve as the foundation for evaluating model performance across diagnostic tasks. Their summary statistics are in Table~\ref{tab:statistics}. We randomly sample up to 1000 instances from each dataset for evaluation.

While foundational, existing datasets for children's speech-language pathology research are largely confined to transcription and binary classification of speech as either disordered or typical. This scope is critically insufficient for developing and validating models for more nuanced clinical applications, such as identifying specific disorder types or their underlying symptoms—categories for which no large-scale, publicly available datasets currently exist. This lack of fine-grained annotation represents a significant bottleneck, impeding the development of automated tools that can support SLPs in differential diagnosis and therapy planning. To address this foundational gap, we collaborated closely with a certified SLP professional to develop a detailed annotation schema that captures both disorder types and their characteristic symptoms. For each speech sample, we assign the most prominent disorder type and symptom, prioritizing the most salient diagnostic features when multiple conditions may co-occur. Speech samples exhibiting no observable signs of speech disorder are annotated as typical. Our annotation protocol and chosen taxonomy are informed by clinical guidelines from the U.S. National Institutes of Health~\cite{simon2016speech} and SLP best practices~\cite{asha_speech_sound_disorders}. After initial manual labeling by our team, we conducted a verification phase in which the speech-language pathologist reviewed annotations to ensure consistency and clinical validity.

We evaluate five core tasks that collectively capture the essential stages of pediatric SLP, from initial screening to detailed diagnostic analysis: \textbf{(1) Disorder Diagnosis}, which assesses a model's ability to distinguish between typical and disordered speech—a critical early triage step for prioritizing clinical resources; \textbf{(2) Transcript-based Diagnosis,} which serves as a baseline for diagnostic accuracy by testing the assumption that speech from children with disorders deviates from expected utterances. This approach operates by matching model-generated transcripts to required spoken text, which offers a minimal, interpretation-free method that could be readily deployed in clinical settings. By benchmarking against this heuristic, we quantify the value added by more sophisticated multimodal LLMs; \textbf{(3) Transcription}, which measures the fidelity of automatic speech recognition (ASR) systems on child with disordered speech, a prerequisite for downstream diagnostic and documentation tasks; \textbf{(4) Disorder Type Classification}, which probes whether models can differentiate between \textit{articulation disorders}—motor-based speech errors such as lisps—and \textit{phonological disorders}, which involve rule-based sound pattern errors such as omission of a part of a word shape (e.g., ca/cat) or consistent shifts in the place of production (e.g., /k/ $\to$ /t/). For example, a child who consistently replaces final consonants, such as saying ``gape'' for ``gate,'' is exhibiting a pattern-based phonological process. A successful model must identify this underlying rule, distinguishing it from an articulation disorder where a child might struggle to physically produce the /t/ sound in any context; \textbf{(5) Disorder Symptom Classification}, a more granular task, requires models to pinpoint the specific clinical symptoms that constitute these error patterns. This includes identifying substitutions, such as the final consonant changing in the production of ``gape'' for ``gate'', or omissions, where a child might say ``gore'' instead of ``gorge'' by dropping the final sound. The task also includes identifying additions (the insertion of extra sounds) and stuttering (disruptions in speech fluency). Accurate identification of these individual symptoms is a critical prerequisite for downstream diagnostic analysis and informs the design of targeted therapy plans. Figure~\ref{fig:taxonomy} illustrates an overview of classification tasks in our pipeline. Details prompts are presented in Appendix~\ref{appendix-prompt}. 

Evaluation metrics for classification tasks include Macro F1, Micro F1, and Exact Match Accuracy, while transcription performance is assessed using Word Error Rate (WER), Match Error Rate (MER), and Word Information Preserved (WIP). The Macro F1 and Micro F1 scores were specifically chosen to provide a comprehensive assessment of classification performance. Macro F1 score showcases the model's performance on each class equally weighted, ensuring that performance on rare but clinically significant disorder categories is not overlooked. In contrast, the Micro F1 score aggregates performance across all individual samples, offering a measure of overall classification correctness.

We implement two distinct model inference pipelines. The first, referred to as the audio-to-LLM prompting pipeline, is designed for models with native multimodal capabilities (e.g., GPT-4o-Audio, Gemini 2.0 Flash). Here, raw audio inputs are passed directly to the model alongside a task-specific prompt. The second pipeline, termed transcription-based prompting, targets language-only models (denoted with the -transcribe suffix). Here, audio inputs are first transcribed using a base automatic speech recognition model (e.g., Whisper or GPT-4's internal ASR), and the resulting text is embedded into a structured prompt for downstream reasoning. This two-pronged architecture compares models with native audio understanding to those using ASR-to-LLM pipelines, highlighting trade-offs between direct and transcription-mediated processing.

\begin{figure*}[!htbp]
  \centering
  \includegraphics[width=\textwidth]{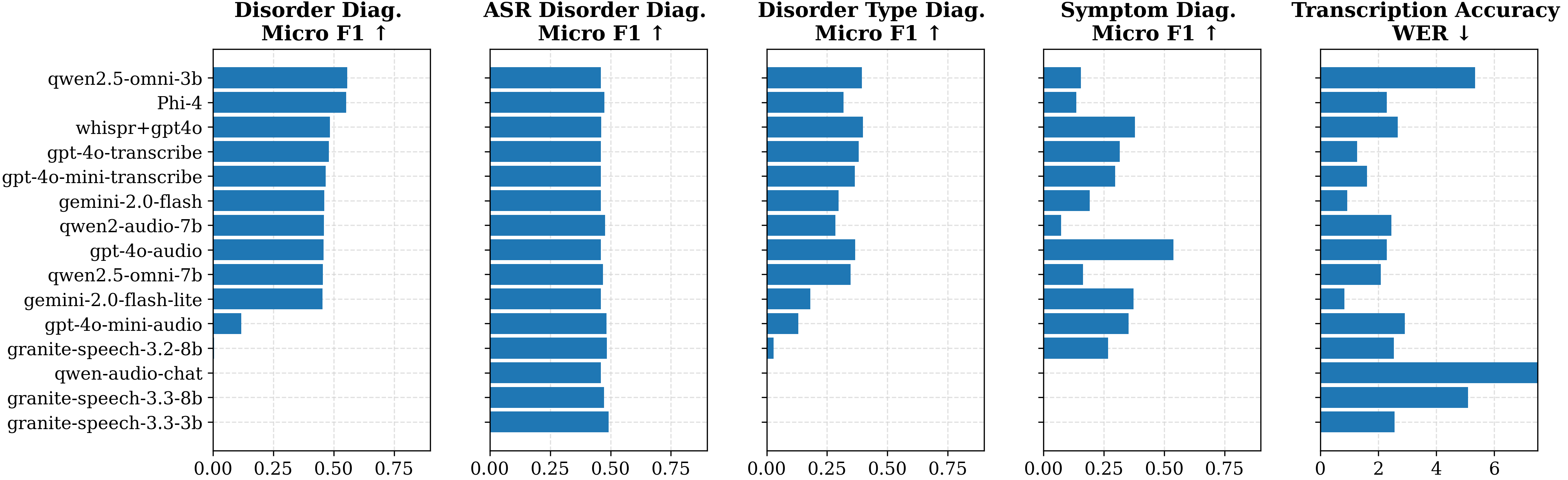} 
  \caption{Metrics across all scenarios}
  \label{fig:metrics}
\end{figure*}

\subsection{Models and Finetuning}
We evaluate 15 speech models. Among the closed-source LLMs, we study GPT-4 family (4o-audio, 4o-mini-audio, 4o-transcribe, and 4o-mini-transcribe), Whisper, and the Gemini 2.0 family (2.0-flash, 2.0-flash-lite). For open-source models, we study Qwen families (2.5-omni-7b, 2.5-omni-3b, 2-audio-7b, audio-chat), the Phi-4, and IBM Granite series (3.3-8b, 3.3-3b, 3.2-8b). These models were chosen to cover a range of model sizes and families across both closed and open sources. 

To investigate the impact of fine-tuning on model performance across multiple tasks, we explore two fine-tuning strategies. Our first strategy involves fine-tuning the model on a speech recognition task (Scenario 3, as described above), relying on the model’s intrinsic ability to transfer knowledge to improve performance on related tasks. In this setup, both typical and disordered speech samples are labeled with the same expected transcriptions. However, assigning identical transcriptions to acoustically distinct inputs may introduce ambiguity and limit the model's ability to learn disorder-specific patterns. To mitigate this, our second strategy - Finetuned w/ Markers - modifies the labeling of disordered speech by appending an asterisk to each word in its transcription. This lightweight labeling scheme serves to differentiate disordered speech from typical speech, thereby guiding the model to better recognize and transcribe disordered speech patterns without altering the overall task formulation.  Details of fine-tuning prompts and hyperparameters are presented in Appendix~\ref{sec:finetuning_hparams}.

We hypothesize that fine-tuning on a general task (e.g., speech recognition) alone is insufficient to yield improvements on specialized clinical tasks unless the fine-tuning data contains explicit information relevant to those tasks. This stems from the theoretical premise that general-purpose models primarily optimize for surface-level acoustic-linguistic alignment, which may not encode the deeper, disorder-specific features, such as atypical phonological patterns or motor-based distortions, necessary for clinical inference~\citep{shor19_interspeech,dorfner2024biomedical}.

%% file: sections/4result.tex
\section{Experiment}
\subsection{Per Scenario Results}
Our findings suggest that both existing proprietary and open-source models currently fail to meet clinically acceptable performance thresholds. This limitation is likely attributable to the underrepresentation of disordered speech in training corpora, as such data is significantly less prevalent than typical speech samples available online. Although F1 scores are not consistently reported in regulatory filings for FDA-approved AI diagnostic systems, published validation studies and clinical evaluations indicate that when F1 is reported, it typically lies within the range of approximately 0.80 to 0.96~\citep{rava2021validation,bagcilar2023lvo}, which serves as a practical standard for clinical viability. Furthermore, model performance varies across different task scenarios, highlighting the absence of a universally robust model that can consistently address the diverse requirements of pediatric SLP applications. Figure~\ref{fig:metrics} presents an overview of the performance of all models. 

\paragraph{Scenario 1: Disorder Diagnosis} In the disorder diagnosis task, performance remains limited, with no model exceeding a micro F1 score of 0.56. The best result is achieved by Qwen-2.5-Omni-3B. Smaller variants within each family exhibit similar performance. Audio-grounded Granite models perform poorly (F1 < 0.1), likely due to their pretraining focus on speech-to-text and translation tasks~\citep{saon2025granite_speech}.

\paragraph{Scenario 2: Transcription-based Diagnosis}
In this scenario, we evaluate a naive baseline that performs diagnosis on transcribed text rather than directly from audio. Figure~\ref{fig:metrics} interestingly shows that all models achieve very similar performance under this setup. This uniformity likely stems from the rigid nature of the scoring criteria: if the ASR output differs even slightly from the expected prompt, the system classifies the speech as disordered. Because around half of the samples in the dataset come from children with speech disorders, this simplistic ``transcribe-and-compare'' approach frequently produces the expected diagnostic label, inflating micro F1 scores. However, the method tends to over-predict disorder and struggles with false positives, which is reflected in consistently lower macro F1 as shown in \ref{tab:detailed_results}. Thus, while micro F1 scores may superficially suggest competitive performance, the underlying strategy is brittle and ultimately not well-suited for robust disorder diagnosis.

\paragraph{Scenario 3: Transcription}
WER varies widely, from 0.83 to 9+, with the Gemini and OpenAI family of models performing the best. Importantly, transcription fidelity shows limited correlation with diagnostic accuracy: Qwen2.5-Omni-3b records a poor WER of 5.3, yet is consistently one of the strongest in classification tasks. These findings indicate that high-quality transcripts are neither necessary nor sufficient for dependable clinical reasoning.

\paragraph{Scenario 4: Disorder Type Classification} Here, Whisper-GPT4o outperforms all models regardless of access or scale, hinting at architectural or pretraining advantages. The transcript-based inference consistently outperforms its audio-grounded counterparts. While for the larger GPT-4o model, audio-grounded inference performs close to its transcript-based counterpart, this trend reverses dramatically for the smaller 4o-mini model, where the ASR+LLM pipeline significantly outperforms the audio-grounded inference. This suggests that while large models can leverage the rich signals in raw audio, smaller models benefit substantially from the structured and potentially less noisy input of a transcript. 

\paragraph{Scenario 5: Disorder Symptom Classification} Accurate identification of these symptoms directly informs treatment goals and therapy design in speech-language pathology. GPT-4o leads in performance but still falls well short of clinically actionable accuracy. Moreover, transcription-based models underperform across all metrics, underscoring that critical acoustic cues needed for symptom detection are often lost or degraded during transcription.

\subsection{Robustness Analysis}
\paragraph{The Effect of Finetuning}
\begin{figure*}[htbp]
\centering
  \includegraphics[width=1.0\textwidth]
  {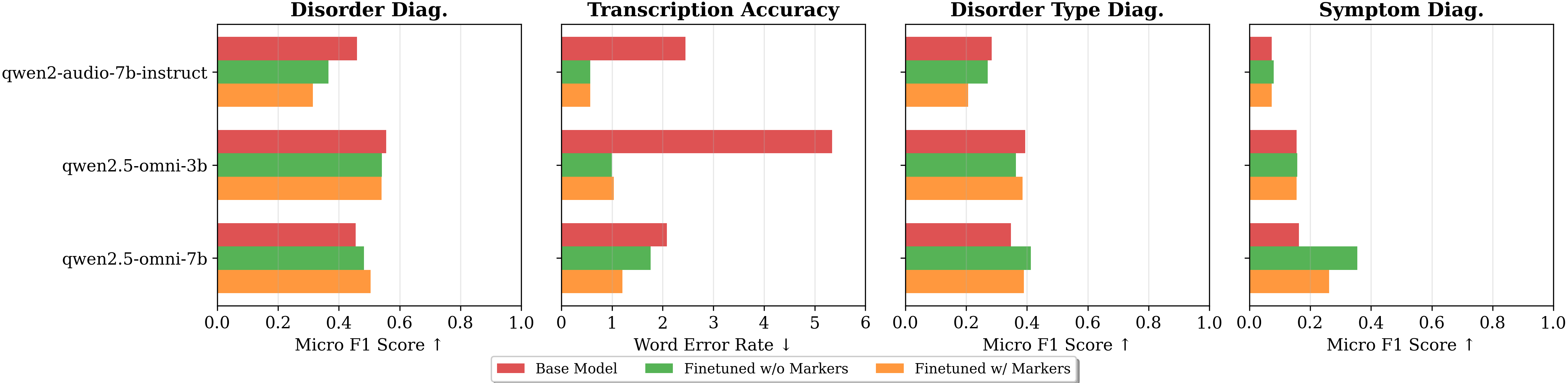} 
  \caption{Model performance after finetuning}
  \label{fig:fine-tuning_performance}
\end{figure*}

\begin{figure*}[htbp]
\centering
  \includegraphics[width=1.0\textwidth]
  {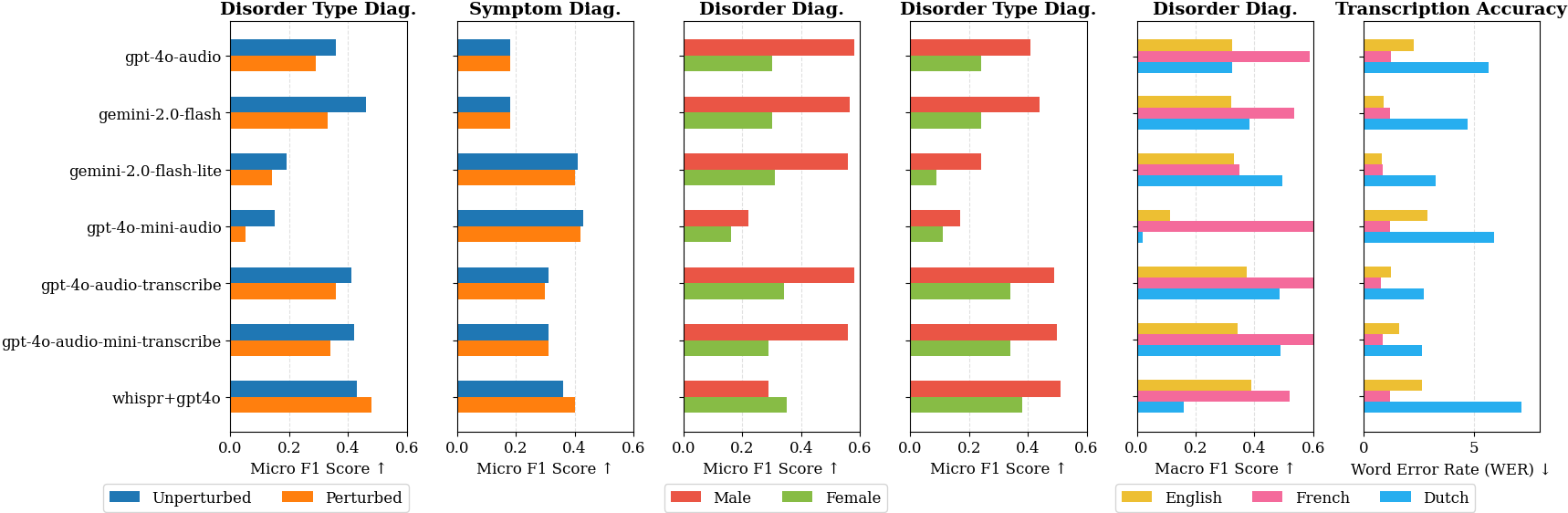} 
  \caption{Model performance across robustness under noisy conditions, across gender, and languages}
  \label{fig:result_analysis}
\end{figure*}

Fine-tuning large models can significantly enhance their performance on downstream tasks. Fine-tuning solely on automatic speech recognition data, regardless of whether disordered speech is explicitly marked, leads to noticeable improvements in ASR-based tasks (Scenarios 2 and 3). Unfortunately, the improvements across other scenarios are not consistent among models. Notably, Qwen2.5-Omni-7b also shows gains in Scenarios 4 and 5, whereas the other models do not. Interestingly, in Scenario 2, models trained with an asterisk achieve better results, likely because the training data help it distinguish more effectively between disordered and typically developing children, though this advantage does not extend to the other scenarios. 

\paragraph{The Effect of Gender} \autoref{fig:result_analysis} showcases that across two diagnostic tasks, the models exhibit a systematic gender performance gap that favors male speech. For both tasks, we run model evaluation on 1000 utterances for each gender on the UltraSuite dataset since it makes demographic identifiers available through its metadata. The results showcase a remarkably consistent pattern: almost every model posts a positive male-female differential. Notably, the gap is not confined to a particular modeling strategy; it appears in fully audio-grounded systems as well as in transcript-conditioned variants. Persistent speech-sound disorders are more commonly associated with boys than girls~\citep{Wren2016}, which can lead to an imbalance in training data and downstream model performance. The magnitude of the divergence suggests practical consequences for clinical deployment and underscores the need for targeted auditing and, potentially, gender-balanced fine-tuning to ensure equitable diagnostic performance across child speakers.

\paragraph{The Effect of Language} \autoref{fig:result_analysis} shows a diverging pattern: classification accuracy shows no clear trend between languages, yet WER in Dutch is markedly worse compared to English and French. A plausible explanation lies in the way these systems were pre-trained. Their ASR components are heavily optimized on English text-to-speech pairs, so lexical recognition degrades when confronted with French or Dutch phonotactics, inflating WER. This effect is likely amplified by the disproportionate number of speakers across these languages - English and French have substantially larger speaker populations than Dutch, which in turn affects the amount of data available during pre-training and leads to poorer performance in lower-resource languages like Dutch. By contrast, the diagnostic classifiers operate on higher-level acoustic embeddings learned during large-scale audio pre-training that are more likely to be language-agnostic~\citep{klempir2024wav2vec}. Those embeddings could still capture phonological and articulatory cues relevant to speech-disorder detection, so classification accuracy can be divergent from word-level transcription accuracy. This lack of correlation between the diagnostic capabilities of a model and its performance under transcription tasks highlights that while the classification performance of some models generalizes across languages, transcription is highly language-dependent.

In a preliminary analysis to evaluate model performance on tonal languages, we assessed disorder diagnosis capabilities on datasets of Taiwanese~\cite{Tsay2007} and Cantonese~\cite{Edwards2008} speech. The results indicate a significant failure of current models to generalize to these linguistic contexts. As shown in Figure~\ref{fig:tonal_languages}, performance is markedly degraded, with the best-performing model achieving a Micro F1 score below 0.10 in Taiwanese, while performance in Cantonese approached zero across all models.

Given that these evaluation datasets consisted entirely of speech from typically developing children, the low F1 scores reveal that the models systematically misdiagnosed the vast majority of samples as disordered. This suggests that the acoustic features these models rely on for disorder detection may be conflating the inherent tonal variations of these languages with pathological speech patterns. These findings underscore a critical limitation, indicating that current architectures are unable to accurately comprehend low-resource tonal languages within the SLP context.

\begin{figure}[h!]
  \centering
  \includegraphics[width=\columnwidth]
  {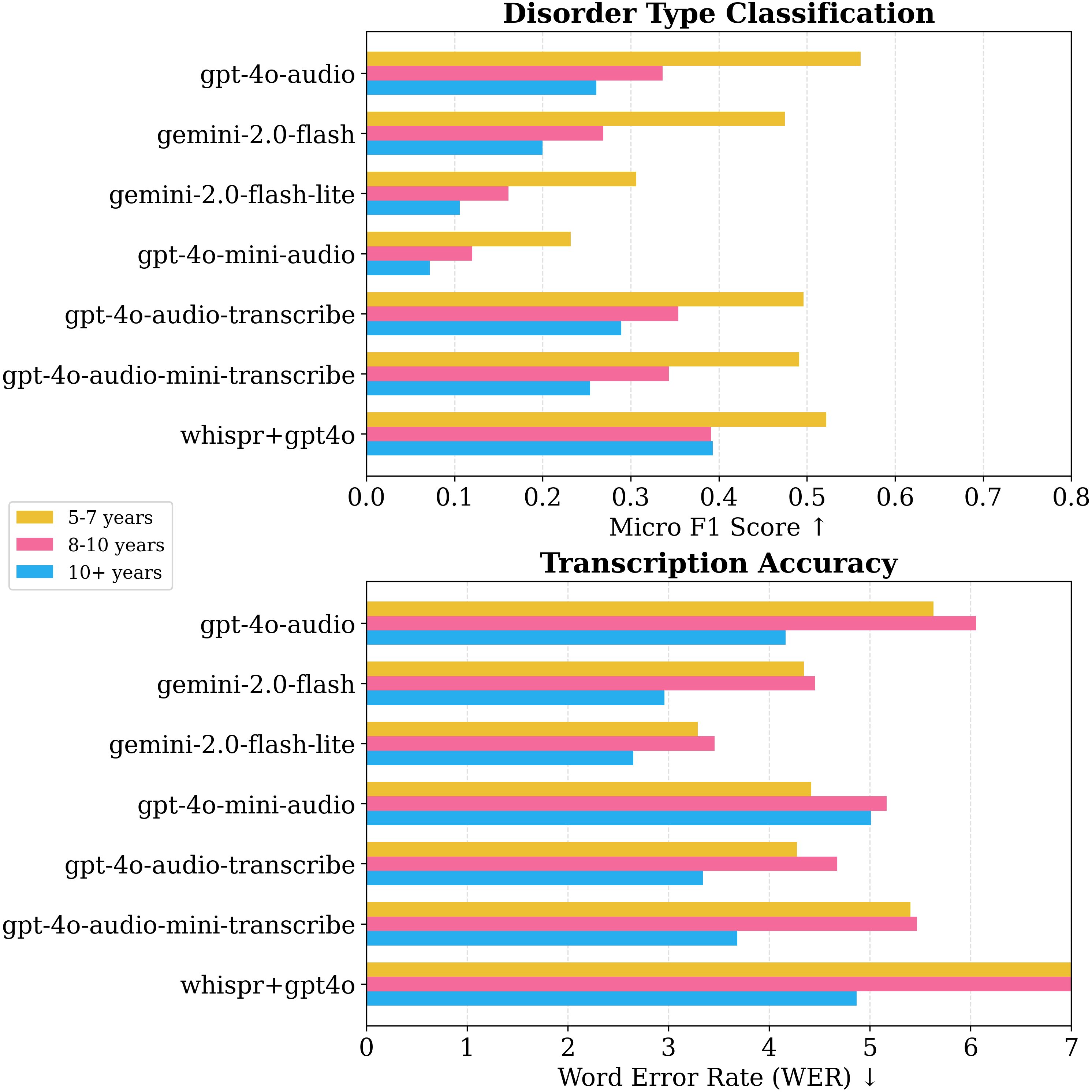} 
  \caption{Model performance across age groups}
  \label{fig:age}
\end{figure}

\paragraph{The Effect of Age} We analyzed model performance across three age ranges: early-elementary (5-7 years old), mid-elementary (8-10 years old), and post-elementary (10+ years old). The UtraSuite dataset provided structured age data for the children in its dataset and was used to run this analysis. The results reveal a distinct trend in disorder type classification. Classification performance is best for younger children and decreases lineraly for older age groups. For transcription accuracy, nearly all models performed best on the speech of the oldest children (10+). Performance gradually decreased for the younger age brackets.

The results reveal that variation in age does not uniformly hinder model performance. One plausible explanation for the inverse relationship between classification performance and age is that younger children’s speech tends to exhibit more salient or exaggerated manifestations of speech-disorder symptoms, making them easier to detect from acoustic features, even when the models are not explicitly tuned for pediatric speech. On the other hand,  given that children's speech gets closer to adult speech with age, transcription performance also improves since the models are pre-trained largely on adult speech-to-text data.

\begin{figure}[htbp]
  \centering
  \includegraphics[width=\columnwidth]
  {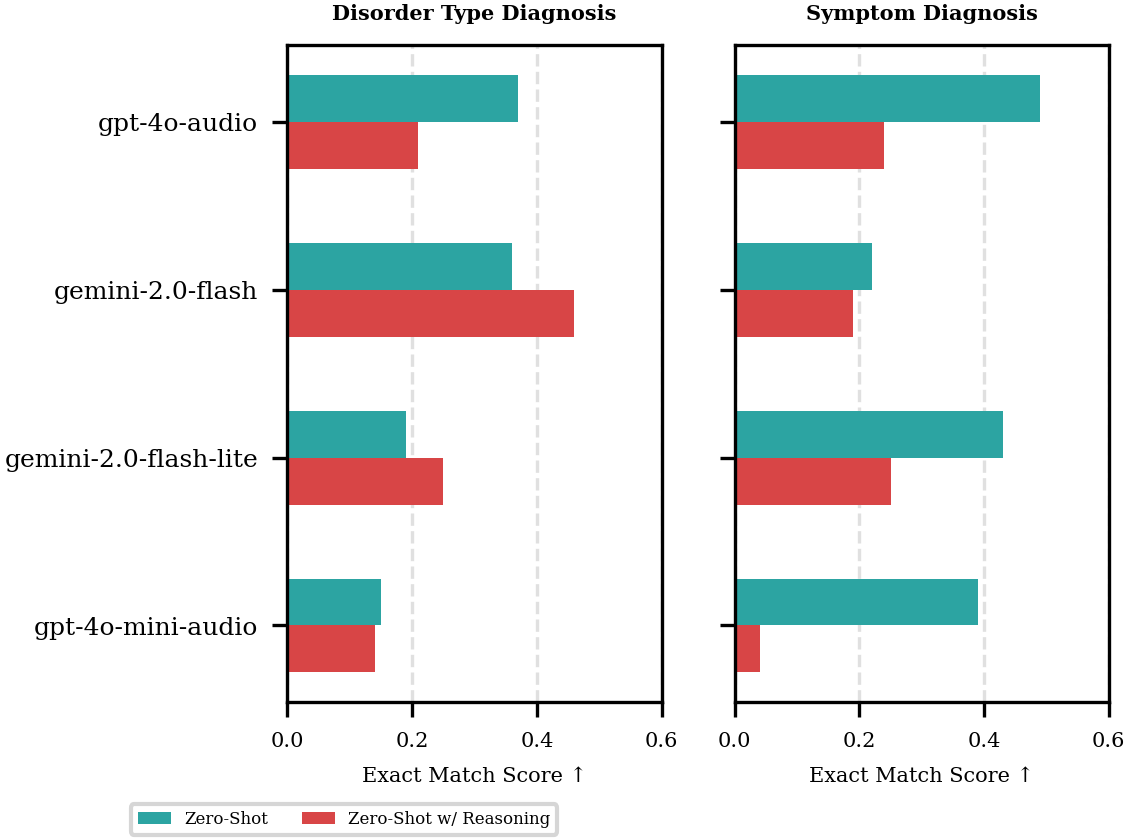} 
  \caption{Compares model's classification performance with/without CoT prompting}
  \label{fig:reasoning_comp}
\end{figure}

\paragraph{The Effect of Reasoning} We analysed the effects of introducing an explicit chain-of-thought (CoT) prompt across the Gemini 2.0 and GPT 4o model families. We saw that CoT systematically depressed F1 scores on the symptom diagnosis task and produced a mixed picture on the disorder type diagnosis, as can be seen in Figure \ref{fig:reasoning_comp}. The pattern aligns with recent evidence that CoT can hamper tasks where the optimal decision boundary is compact or where answer formatting is unforgiving, because the additional reasoning tokens introduce distraction or bleed into the predicted label~\citep{liu2024mind}. However, even when CoT degrades final accuracy, the explicit rationale provides crucial insight into a model's failure modes. By inspecting the reasoning trace, we can diagnose why a model arrived at its conclusion, rather than simply observing the error itself. We analysed 200 reasoning traces, each for the 4 models for the Disorder-Type scenario. The CoTs are analysed along the following axes to reveal a clear profile of the model's reasoning process~\citep{Gandhi2025-fj}. A subset of the analysed traces can be found in Appendix \ref{Reasoning}.

\begin{itemize}
    \item \textbf{Subgoal Setting}, where the model breaks down a complex problem into manageable steps (e.g., ``We first need to...'').
    \item \textbf{Rule following} or the application of taxonomical rules during classification (e.g., repeated errors in the last consonant indicate a phonological disorder rather than articulation).
    \item \textbf{Error Detection} or whether the model can recognize individual errors in speech, on which it will determine its classification.
\end{itemize}

\begin{figure}[htbp]
  \centering
  \includegraphics[width=\columnwidth]
  {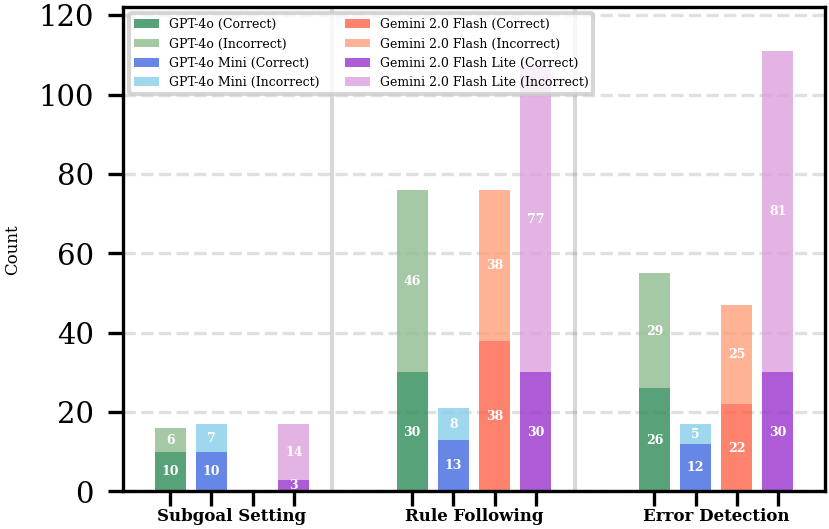} 
  \caption{CoT Behaviour Analysis - Breakdown of behaviors showcased by the model. The bar for each model indicates the number of times the given behaviour is observed. Each bar is broken down by the times the given behaviour resulted in a correct/incorrect answer.}
  \label{fig:reasoning_analysis}
\end{figure}

As shown in Figure \ref{fig:reasoning_analysis}, our analysis of CoT reasoning reveals key differences in the models' problem-solving approaches. Rule Following and Error Detection were the most prominent strategies employed across all models. However, the effective application of these strategies varied significantly. For GPT-4o, a majority of reasoning traces using Rule Following and Error Detection led to an incorrect final answer. This high rate of flawed application likely explains its reduced classification performance with CoT. Conversely, Gemini 2.0 Flash's improved performance correlates with more reliable reasoning; its application of Rule Following, for example, was correct in 50\% of instances, compared to 39.5\% for GPT-4o. This suggests that the fidelity of a model's reasoning directly impacts its accuracy. Interestingly, Gemini 2.0 Flash Lite exhibits the highest frequency of Rule Following and Error Detection behaviors but with the lowest accuracy (28\% and 27\%, respectively). Despite this, its classification performance improves, highlighting that the frequency of these model behaviours also contributes to overall model performance. This analysis indicates that for SLP tasks, CoT prompting improves performance only when the label space is limited with well-defined boundaries and the model can reliably apply rule-following and error-detection strategies in its reasoning traces.

\paragraph{Effect of Ensemble}
We implemented an ensemble strategy that combines predictions from multiple models. For classification tasks, the ensemble uses a majority voting system where the final prediction is determined by selecting the most frequently predicted label across all models, effectively leveraging the collective wisdom of multiple classifiers to reduce individual model biases and improve robustness. We created ensembles of three models to avoid tie scenarios where possible. If encountered, a tie is broken by arbitrarily selecting one of the predicted labels. 

The ensemble analysis in Table \ref{tab:detailed_results} reveals a nuanced performance landscape. For the primary task of disorder diagnosis, a performance plateau is observed, with three distinct ensembles, the ``Qwen Pure Ensemble'' and both ``Google + OpenAI'' combinations, achieving similar F1 scores. However, significant differences emerge in the more granular classification tasks. The mixed-vendor Google + OpenAI ensembles demonstrate substantially better performance in symptom identification, achieving a Micro F1 score of around 0.38, more than double that of the Qwen-based ensembles. Conversely, the Qwen ensembles show a slight advantage in disorder type classification. These findings suggest that a simple ensemble strategy does not guarantee superior performance, and that single model family ensembles are not guaranteed to be dominant. The effectiveness of an ensemble appears to be highly task-dependent and may benefit from combining models with diverse architectures, as seen with the Google and OpenAI ensembles' success in symptom classification. This highlights that the optimal ensemble configuration is not merely a combination of the strongest individual models but requires empirical testing to find complementary strengths for specific tasks.

\paragraph{The Effect of Background Noise} We analyze model robustness by evaluating their performance with three added artificial perturbations, including road noise, classroom noise, and office noise. We aggregate the results to assess model performance. Each type of background noise was added at 20dB. As shown in \autoref{fig:result_analysis}, model performance degrades on average by 10 basis points across all models for disorder type diagnosis, while performance on symptom diagnosis remains virtually unchanged. These observations suggest that noise resilience is not strictly determined by model and inference architecture. It might instead be influenced by model-specific factors such as design, scale, and training data. 

%% file: sections/5discussion.tex
\section{Conclusion \& Future Work}
Our empirical findings reveal that even the best-performing models fall short of clinical-grade reliability, revealing considerable room for improvement. This performance gap presents a significant clinical risk, as the deployment of an unreliable model could lead to patient misdiagnosis or delayed treatment. Our fine-tuning experiments with the Qwen2.5 family demonstrate that this performance can be substantially improved, highlighting the effectiveness of task-specific adaptation and the potential for developing specialized SLP models that generalize well across tasks.

Our robustness analysis identifies a consistent performance disparity favoring male speakers, contradicting the clinical principle of equitable care and highlighting the need for bias mitigation through strategies such as gender-balanced fine-tuning and targeted data augmentation. Cross-linguistic evaluations further demonstrate that audio-grounded models retain competitive diagnostic performance even when transcription quality degrades, suggesting that higher-order acoustic features support language-agnostic reasoning.

Future work could expand coverage to low-resource languages and neurodiverse populations, and evaluate model explanations for clinical faithfulness. We also plan to investigate privacy-preserving fine-tuning paradigms to facilitate deployment in sensitive pediatric settings. Collectively, these directions aim to bridge the gap between promising laboratory advances and the development of clinically robust, ethically sound AI systems for SLP.

\section*{Limitations and Ethical Considerations}

Despite the comprehensive scope of our benchmark, several limitations and core ethical considerations warrant explicit discussion prior to any consideration of clinical deployment.

\subsection*{1. Privacy and Data Sensitivity}
The ethical considerations surrounding privacy and consent are central to work in sensitive clinical domains. While this study was conducted on public datasets collected with informed consent and subsequently anonymized to protect participant privacy, our current setup does not incorporate a formal privacy-preserving learning or evaluation framework. Addressing this is essential to safeguard patient data and build trust among clinicians and families. As such, future work must investigate privacy-preserving fine-tuning paradigms to facilitate responsible deployment in pediatric settings.

\subsection*{2. Bias and Equity}
A primary finding of this work is the identification of significant performance disparities, which present a direct challenge to the principle of equitable care.
\begin{itemize}
    \item Gender Bias: Our evaluation revealed a consistent performance disparity that favors male speakers across multiple models. This indicates an urgent need for bias mitigation strategies, such as gender-balanced fine-tuning and targeted data augmentation, which were not implemented in this study but are critical next steps.
    \item Data Representation: The datasets employed are drawn primarily from English and French speakers, leading to an underrepresentation of other languages and dialects. This limits the generalizability of our findings and risks creating tools that are not effective for more linguistically and culturally diverse populations.
    \item Annotation Bias: The annotation process, though guided by an SLP professional, required prioritizing the most prominent disorder when multiple conditions co-occurred, introducing a degree of subjectivity.
\end{itemize}

\subsection*{3. Clinical Reliability and Validation}
A core ethical requirement for deployment is robust clinical validation.
\begin{itemize}
    \item Performance Gap: This study establishes a benchmark for clinical viability based on existing FDA-approved systems (F1 scores of 0.80 to 0.85). Our findings demonstrate that even the best-performing models currently fall short of this standard.
    \item Risk of Misdiagnosis: This performance gap represents a significant clinical risk, as the deployment of an unreliable model could lead to patient misdiagnosis or delayed treatment. The results underscore that current MLMs are not yet clinically robust and require substantial further development and validation before they can be considered for supportive roles in real-world SLP workflows.
\end{itemize}

%% file: sections/7ack.tex
\section*{Acknowledgments}
SK acknowledges support by NSF 2046795 and 2205329, IES R305C240046, ARPA-H, the MacArthur Foundation, Schmidt Sciences, Stanford HAI, and Google. NH acknowledges the National AI Institute for Exceptional Education (Institute of Education Sciences, U.S. Department of Education, through Grant 22298673 (NSF)).

%% file: sections/6appendix.tex
\section {Dataset Statistic}

\begin{table}[hbt!]
    \centering
    \caption{Dataset statistics}
    \label{tab:statistics}
    \resizebox{0.5\textwidth}{!}{
    \begin{tabular}{lccccc}
        \toprule
        \textbf{Dataset} & \textbf{\# Children} & \textbf{\# Samples} & \textbf{Age Range}\\
        \midrule
        Ultrasuite & 66 & 8338 & 5--13 \\
        ENNI & 377 & 16546 & 4--9\\
        LeNormand (French) & 17 & 329 & 3--8\\
        PERCEPT-GFTA & 350 & 3664 & 6--17\\
        \bottomrule
    \end{tabular}
    }
\end{table}

\section {Additional Related Works}
\paragraph{AI in Speech Language Pathology Assessment} The use of LLMs in clinical speech-language assessment has gained increasing attention in recent years. Several recent studies have demonstrated the utility of LLMs in detecting and characterizing speech and language disorders. For instance, \citet{bhattacharya2024aphasia} showed that pre-trained LLMs could effectively identify both the presence and type of aphasia, suggesting that these models can serve as viable tools for clinical screening and diagnosis of language disorders. Beyond perception studies, a growing body of technical literature examines the use of speech and language features for automated assessment. \citet{engelhardt2021review} reviewed computational features used to assess cognitive and thought disorders, highlighting the relevance of acoustic and linguistic cues in differential diagnosis. Similarly, \citet{heilmann2023automation} demonstrated that automatic language sample analysis tools can support clinical workflows, providing reliable linguistic metrics with reduced human effort. 

\paragraph{LLMs for Disordered Speech Analysis} A recent survey of SLPs and graduate students revealed a combination of cautious optimism and skepticism regarding the integration of LLMs such as ChatGPT into diagnostic and therapeutic workflows~\cite{austin2024perceptions}. These practitioner attitudes highlight critical socio-technical barriers to the clinical adoption of AI-driven systems in speech-language pathology. Recent research has explored the adaptation of LLMs for disordered speech processing.\citet{sanguedolce2024universal} proposed a more generalized framework by fine-tuning Whisper on a dataset of stroke patients, resulting in a universal disordered-speech detection model. Their approach exhibited strong generalization across multiple neurological conditions, underscoring the potential of foundation models for broad-spectrum clinical speech applications.

\section{Additional Analysis}
\begin{figure}[htbp]
\centering
  \includegraphics[width=\columnwidth]
  {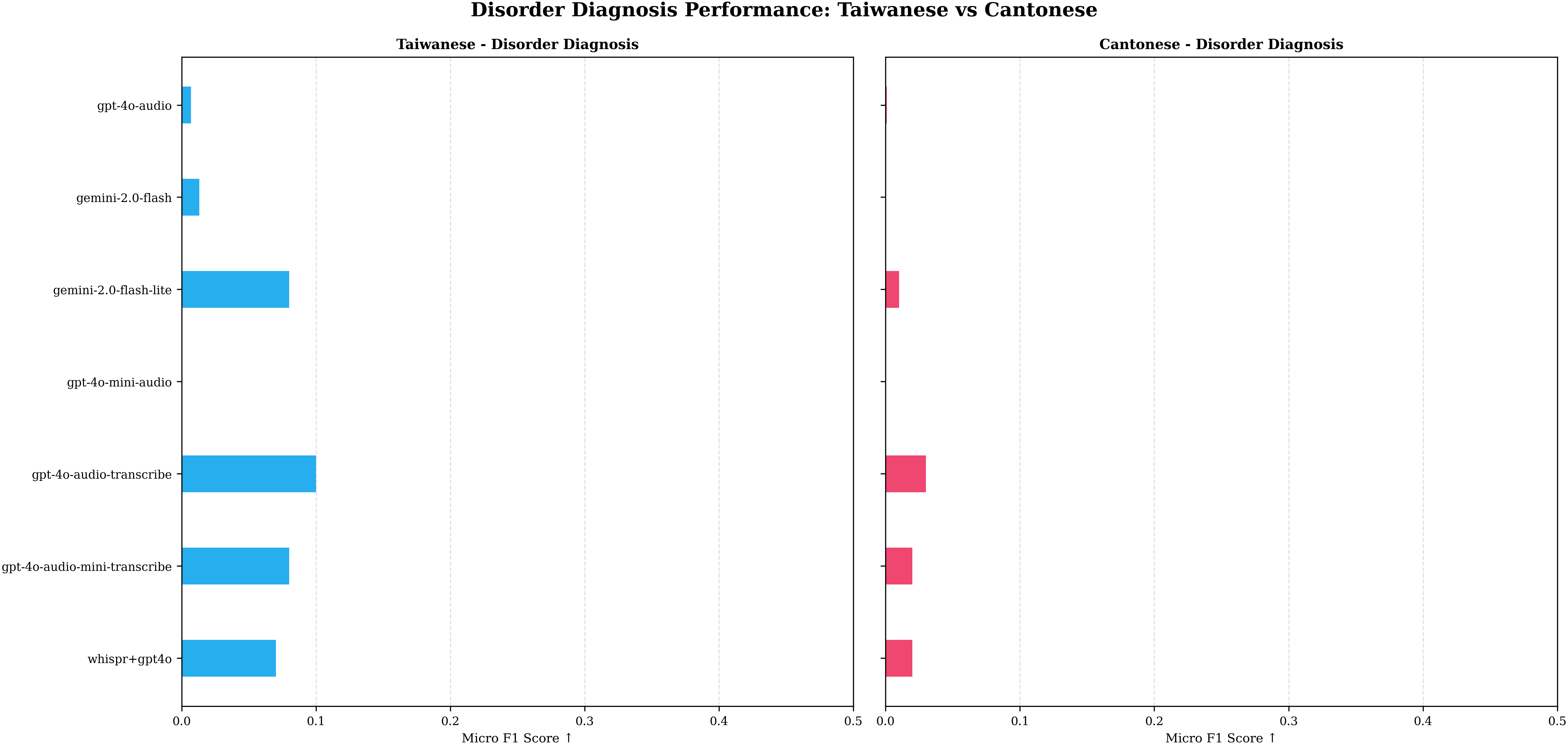} 
  \caption{Disorder Diagnosis performance on tonal languages}
  \label{fig:tonal_languages}
\end{figure}

\paragraph{The Effect of Fewshot Examples} The results of the GPT-4 family across the first three scenarios under few-shot prompting indicate that few-shot examples do not consistently enhance the model's intrinsic capabilities; the benefits of prompting are not uniformly evident. For instance, while few-shot prompting significantly improves the performance of GPT-4o-Mini-Transcribe and GPT-4o-Transcribe in Scenario 1, it leads to a reduction in accuracy in Scenario 2. Given that the examples provided to the model describe the expected transcription, few-shot prompting may bias the model to transcribe what the child was expected to say rather than accurately transcribe what the child is saying, which would explain the reduced diagnostic accuracy in Scenario 2. Our observations are consistent with prior findings on text-only LLMs ~\citep{jacovi-etal-2023-comprehensive}.

\section{Fine-tuning details}
\label{sec:finetuning_hparams}

\subsection{Single Finetuning Round}
We perform supervised fine-tuning on three models, including Qwen2-Audio 7B, Qwen2.5-Omni 3B, and Qwen2.5-Omni 7B using LLaMA-Factory framework~\cite{zheng2024llamafactory}. We reuse the same UltraSuite dataset, but filtered out children whose speech was in the evaluation set to avoid data contamination. We set the same fine-tuning hyperparameters across all models. Specifically, we used a LoRA rank of 32 with a LoRA alpha of 64, applying LoRA to all linear layers. The maximum token length was 4096, and training was performed with a batch size of 32 for three epochs. We adopted a learning rate of 0.0001 with cosine scheduling and a warm-up ratio of 0.1. Regarding the prompts used for the three ablation settings for fine-tuning models, we present them as follows.

\begin{enumerate}
    \item ASR-only without asterisk: \texttt{<audio>Transcribe this sound into text.}
    \item ASR-only with asterisk: \texttt{<audio>Transcribe this sound into text. If the speech is disordered, please mark the words with an asterisk.}
\end{enumerate}

\subsection{Multiple Finetuning Rounds}
To investigate the model’s capacity for self-evolution, we conducted an iterative finetuning experiment in which the training data were generated by the model itself. The procedure is summarized as follows. In each iteration, we prompted the current model to generate up to 2,000 samples comprising both normal and disordered text–speech pairs. The details of the data generation process are described below. The generated data were then used to fine-tune the same model, after which the updated model was evaluated using our proposed framework. This procedure was repeated for five iterations. We use the Qwen 2.5 Omni 3B model for this experiment, and the finetuning hyperparameters were kept identical to those used in the single-round finetuning experiment described above.

\paragraph{Data generation process.} We employed the ViLLM framework to deploy the current model and prompt it to produce both disordered and normal text and corresponding speech. We use simple sentences from the Tatoeba dataset\footnote{https://huggingface.co/datasets/Helsinki-NLP/tatoeba\_mt} as the input text. The prompt used to generate text and speech is as below.
\begin{itemize}
    \item Normal text and speech: \texttt{Repeat exactly the following sentence and do not alter it. Return only the sentence and nothing else. The sentence is: \{transcript\}}
    \item Disordered text and speech: \texttt{Assuming you are a child with speech disordered. \{disorder\_description\} Generate a corresponding transcript with the `\{disorder\_type\} disorder` when the child say: "\{transcript\}". Return only the transcript text and nothing else.}
\end{itemize}
If the model failed to generate disordered text, we manually introduced disorder markers into the text (including addition, substitution, omission, and stuttering). We then instructed the model to regenerate the disordered text and corresponding speech using the prompt as above. All generated audio was further processed through a \emph{child-en} pipeline, which applied a series of signal processing operations (including pitch shifting, low-frequency reduction, high-frequency boosting, and slight speed-up) to approximate the characteristics of children’s speech. Subsequently, each sample was verified for detecting disorder using the same current model. If verification failed, generation was repeated up to five times; if it still failed, the sample was discarded, and the process moved on to the next sample. The prompt we used for verification is presented as follows.

\begin{lstlisting}
You are a highly experienced Speech-Language Pathologist (SLP). An audio recording of a child's voice will be provided. The child is attempting to say: "\{transcript\}". Based on your professional expertise, analyze the child's speech production in the recording to determine whether it indicates a potential speech-language disorder (addition, substitution, omission, stuttering). Respond with one of the following labels only: `speech\_disorder`, `typically\_developing`. Do not include any explanation or additional text in your response.}
\end{lstlisting}

We present the results of the self-evolving experiment in Table~\ref{tab:detailed_results}. As shown, the transcription task exhibits a clear reduction in word error rate, which can be attributed to the availability of the ground-truth transcript for each generated speech sample. In contrast, the performance on disorder diagnosis remains largely unchanged across iterations. This is expected, as the same model is used to verify whether the speech is disordered, thereby providing no additional learning signal. Consequently, the model’s performance on this task does not surpass that of the original model. A promising direction for future work is to employ a stronger verification model and/or incorporate supervision on the disorder type, which may yield richer signals for iterative finetuning.

\section{Detailed Results}
\ref{tab:detailed_results} presents our evaluation results. Values are reported as mean ± bootstrap standard deviation, providing both the point estimate and a measure of uncertainty in the metric estimates.

\begin{table}[!htbp]
\centering
\caption{Detailed Model Performance Results}
\label{tab:detailed_results}
\resizebox{0.9\textwidth}{!}{%
\begin{tabular}{l|cccc|cccc|cccc|ccc}
\toprule
& \multicolumn{4}{c|}{\textbf{Micro F1} $\uparrow$} & \multicolumn{4}{c|}{\textbf{Macro F1} $\uparrow$} & \multicolumn{4}{c|}{\textbf{Exact Match} $\uparrow$} & \multicolumn{3}{c}{} \\
\textbf{Model} & \shortstack[c]{Disorder\\Diag.} & \shortstack[c]{ASR\\Diag.} & \shortstack[c]{Disorder\\Type} & \shortstack[c]{Symptom\\Diag.} & \shortstack[c]{Disorder\\Diag.} & \shortstack[c]{ASR\\Diag.} & \shortstack[c]{Disorder\\Type} & \shortstack[c]{Symptom\\Diag.} & \shortstack[c]{Disorder\\Diag.} & \shortstack[c]{ASR\\Diag.} & \shortstack[c]{Disorder\\Type} & \shortstack[c]{Symptom\\Diag.} & \textbf{WER} $\downarrow$ & \textbf{MER} $\downarrow$ & \textbf{WIP} $\uparrow$ \\
\midrule
gemini-2.0-flash-lite & 0.454 $\pm$ 0.015 & 0.4590 $\pm$ 0.011 & 0.180 $\pm$ 0.012 & 0.373 $\pm$ 0.016 & 0.330 $\pm$ 0.010 & 0.3146 $\pm$ 0.012 & 0.174 $\pm$ 0.011 & 0.171 $\pm$ 0.024 & 0.454 $\pm$ 0.015 & 0.4590 $\pm$ 0.011 & 0.180 $\pm$ 0.012 & 0.373 $\pm$ 0.016 & \textbf{0.83 $\pm$ 0.016} & 0.723 $\pm$ 0.016 & 0.215 $\pm$ 0.016 \\
gemini-2.0-flash & 0.461 $\pm$ 0.014 & 0.4600 $\pm$ 0.013 & 0.298 $\pm$ 0.015 & 0.191 $\pm$ 0.013 & 0.322 $\pm$ 0.009 & 0.3151 $\pm$ 0.011 & 0.212 $\pm$ 0.011 & 0.079 $\pm$ 0.005 & 0.461 $\pm$ 0.014 & 0.4600 $\pm$ 0.013 & 0.298 $\pm$ 0.015 & 0.191 $\pm$ 0.013 & 0.931 $\pm$ 0.000 & 0.713 $\pm$ 0.000 & 0.227 $\pm$ 0.000 \\
gpt-4o-mini-audio & 0.117 $\pm$ 0.011 & 0.4838 $\pm$ 0.009 & 0.131 $\pm$ 0.011 & 0.352 $\pm$ 0.017 & 0.113 $\pm$ 0.009 & 0.3711 $\pm$ 0.008 & 0.122 $\pm$ 0.008 & 0.095 $\pm$ 0.005 & 0.117 $\pm$ 0.011 & 0.4838 $\pm$ 0.009 & 0.131 $\pm$ 0.011 & 0.352 $\pm$ 0.017 & 2.92 $\pm$ 0.066 & 0.744 $\pm$ 0.006 & 0.201 $\pm$ 0.006 \\
gpt-4o-audio & 0.458 $\pm$ 0.014 & 0.4590 $\pm$ 0.014 & 0.366 $\pm$ 0.015 & \textbf{0.538 $\pm$ 0.016} & 0.324 $\pm$ 0.005 & 0.3163 $\pm$ 0.015 & 0.207 $\pm$ 0.007 & 0.204 $\pm$ 0.009 & 0.458 $\pm$ 0.014 & 0.4590 $\pm$ 0.014 & 0.366 $\pm$ 0.015 & \textbf{0.538 $\pm$ 0.016} & 2.29 $\pm$ 0.117 & \textbf{0.69 $\pm$ 0.015} & \textbf{0.247 $\pm$ 0.015} \\
gpt-4o-mini-transcribe & 0.466 $\pm$ 0.018 & 0.4600 $\pm$ 0.012 & 0.365 $\pm$ 0.016 & 0.297 $\pm$ 0.016 & 0.344 $\pm$ 0.012 & 0.3168 $\pm$ 0.013 & 0.284 $\pm$ 0.015 & 0.202 $\pm$ 0.014 & 0.466 $\pm$ 0.018 & 0.4600 $\pm$ 0.012 & 0.365 $\pm$ 0.016 & 0.297 $\pm$ 0.016 & 1.61 $\pm$ 0.015 & 0.76 $\pm$ 0.015 & 0.186 $\pm$ 0.015 \\
gpt-4o-transcribe & 0.480 $\pm$ 0.017 & 0.4600 $\pm$ 0.011 & 0.381 $\pm$ 0.016 & 0.316 $\pm$ 0.015 & 0.373 $\pm$ 0.014 & 0.3202 $\pm$ 0.010 & \textbf{0.318 $\pm$ 0.017} & 0.201 $\pm$ 0.019 & 0.480 $\pm$ 0.017 & 0.4600 $\pm$ 0.011 & 0.381 $\pm$ 0.016 & 0.316 $\pm$ 0.015 & 1.267 $\pm$ 0.000 & 0.761 $\pm$ 0.000 & 0.192 $\pm$ 0.000 \\
whisper-gpt4o & 0.484 $\pm$ 0.015 & 0.4611 $\pm$ 0.015 & \textbf{0.399 $\pm$ 0.016} & 0.378 $\pm$ 0.016 & 0.391 $\pm$ 0.013 & 0.3190 $\pm$ 0.014 & 0.245 $\pm$ 0.013 & \textbf{0.222 $\pm$ 0.015} & 0.484 $\pm$ 0.015 & 0.4611 $\pm$ 0.015 & \textbf{0.399 $\pm$ 0.016} & 0.378 $\pm$ 0.016 & 2.668 $\pm$ 0.018 & 0.741 $\pm$ 0.015 & 0.182 $\pm$ 0.015 \\
granite-speech-3.2-8b & 0.004 $\pm$ 0.016 & 0.484 $\pm$ 0.016 & 0.028 $\pm$ 0.016 & 0.268 $\pm$ 0.016 & 0.004 $\pm$ 0.008 & \textbf{0.392 $\pm$ 0.014} & 0.016 $\pm$ 0.009 & 0.090 $\pm$ 0.009 & 0.004 $\pm$ 0.016 & 0.484 $\pm$ 0.016 & 0.028 $\pm$ 0.016 & 0.268 $\pm$ 0.015 & 2.535 $\pm$ 0.135 & 0.716 $\pm$ 0.012 & 0.161 $\pm$ 0.010 \\
granite-speech-3.3-2b & 0.000 $\pm$ 0.017 & \textbf{0.492 $\pm$ 0.017} & 0.000 $\pm$ 0.007 & 0.000 $\pm$ 0.003 & 0.000 $\pm$ 0.013 & 0.388 $\pm$ 0.014 & 0.000 $\pm$ 0.008 & 0.000 $\pm$ 0.003 & 0.000 $\pm$ 0.016 & \textbf{0.492 $\pm$ 0.016} & 0.000 $\pm$ 0.007 & 0.000 $\pm$ 0.003 & 2.564 $\pm$ 0.158 & 0.705 $\pm$ 0.014 & 0.075 $\pm$ 0.007 \\
granite-speech-3.3-8b & 0.000 $\pm$ 0.015 & 0.473 $\pm$ 0.017 & 0.000 $\pm$ 0.016 & 0.000 $\pm$ 0.009 & 0.000 $\pm$ 0.016 & 0.344 $\pm$ 0.011 & 0.000 $\pm$ 0.011 & 0.000 $\pm$ 0.011 & 0.000 $\pm$ 0.016 & 0.473 $\pm$ 0.017 & 0.000 $\pm$ 0.016 & 0.000 $\pm$ 0.009 & 5.094 $\pm$ 0.229 & 0.971 $\pm$ 0.004 & 0.022 $\pm$ 0.004 \\
Phi-4-multimodal-instruct & 0.552 $\pm$ 0.016 & 0.474 $\pm$ 0.016 & 0.318 $\pm$ 0.015 & 0.135 $\pm$ 0.011 & \textbf{0.535 $\pm$ 0.016} & 0.375 $\pm$ 0.014 & 0.163 $\pm$ 0.006 & 0.118 $\pm$ 0.010 & 0.552 $\pm$ 0.017 & 0.474 $\pm$ 0.016 & 0.318 $\pm$ 0.015 & 0.135 $\pm$ 0.011 & 2.288 $\pm$ 0.149 & 0.735 $\pm$ 0.012 & 0.221 $\pm$ 0.011 \\
qwen-audio-chat & 0.000 $\pm$ 0.016 & 0.460 $\pm$ 0.016 & 0.000 $\pm$ 0.000 & 0.000 $\pm$ 0.000 & 0.000 $\pm$ 0.008 & 0.315 $\pm$ 0.008 & 0.000 $\pm$ 0.000 & 0.000 $\pm$ 0.000 & 0.000 $\pm$ 0.017 & 0.460 $\pm$ 0.017 & 0.000 $\pm$ 0.000 & 0.000 $\pm$ 0.000 & 9.549 $\pm$ 1.176 & 0.895 $\pm$ 0.005 & 0.081 $\pm$ 0.005 \\
qwen2-audio-7b-instruct & 0.460 $\pm$ 0.016 & 0.477 $\pm$ 0.017 & 0.284 $\pm$ 0.015 & 0.073 $\pm$ 0.009 & 0.210 $\pm$ 0.008 & 0.361 $\pm$ 0.012 & 0.182 $\pm$ 0.008 & 0.077 $\pm$ 0.010 & 0.460 $\pm$ 0.016 & 0.477 $\pm$ 0.016 & 0.284 $\pm$ 0.015 & 0.073 $\pm$ 0.009 & 2.449 $\pm$ 0.164 & 0.790 $\pm$ 0.011 & 0.168 $\pm$ 0.010 \\
qwen2.5-omni-3b & \textbf{0.556 $\pm$ 0.016} & 0.459 $\pm$ 0.017 & 0.394 $\pm$ 0.016 & 0.155 $\pm$ 0.012 & 0.401 $\pm$ 0.013 & 0.318 $\pm$ 0.008 & 0.272 $\pm$ 0.015 & 0.097 $\pm$ 0.007 & \textbf{0.556 $\pm$ 0.016} & 0.459 $\pm$ 0.016 & 0.394 $\pm$ 0.016 & 0.155 $\pm$ 0.012 & 5.346 $\pm$ 0.323 & 0.777 $\pm$ 0.011 & 0.187 $\pm$ 0.010 \\
qwen2.5-omni-7b & 0.455 $\pm$ 0.016 & 0.468 $\pm$ 0.016 & 0.347 $\pm$ 0.016 & 0.163 $\pm$ 0.012 & 0.425 $\pm$ 0.016 & 0.371 $\pm$ 0.013 & 0.203 $\pm$ 0.009 & 0.105 $\pm$ 0.010 & 0.455 $\pm$ 0.016 & 0.468 $\pm$ 0.016 & 0.347 $\pm$ 0.016 & 0.163 $\pm$ 0.012 & 2.084 $\pm$ 0.143 & 0.762 $\pm$ 0.011 & 0.192 $\pm$ 0.010 \\
\midrule
\multicolumn{16}{c}{\textbf{Finetuned Models with Asterisk}} \\
\midrule
Qwen2-Audio-7B-Instruct & 0.314 $\pm$ 0.017 & \textbf{0.582 $\pm$ 0.016} & 0.207 $\pm$ 0.013 & 0.073 $\pm$ 0.009 & 0.015 $\pm$ 0.008 & \textbf{0.571 $\pm$ 0.017} & 0.035 $\pm$ 0.008 & 0.006 $\pm$ 0.008 & 0.314 $\pm$ 0.016 & \textbf{0.582 $\pm$ 0.016} & 0.207 $\pm$ 0.013 & 0.073 $\pm$ 0.009 & \textbf{0.574 $\pm$ 0.028} & \textbf{0.420 $\pm$ 0.013} & \textbf{0.506 $\pm$ 0.013} \\
Qwen2.5-Omni-3B & \textbf{0.540 $\pm$ 0.016} & 0.499 $\pm$ 0.017 & 0.386 $\pm$ 0.016 & 0.155 $\pm$ 0.012 & \textbf{0.351 $\pm$ 0.007} & 0.463 $\pm$ 0.016 & \textbf{0.285 $\pm$ 0.015} & 0.062 $\pm$ 0.006 & \textbf{0.540 $\pm$ 0.017} & 0.499 $\pm$ 0.017 & 0.386 $\pm$ 0.016 & 0.155 $\pm$ 0.012 & 1.036 $\pm$ 0.102 & 0.540 $\pm$ 0.013 & 0.383 $\pm$ 0.013 \\
Qwen2.5-Omni-7B & 0.504 $\pm$ 0.017 & 0.486 $\pm$ 0.017 & \textbf{0.390 $\pm$ 0.016} & \textbf{0.262 $\pm$ 0.014} & 0.343 $\pm$ 0.009 & 0.454 $\pm$ 0.016 & 0.267 $\pm$ 0.011 & \textbf{0.122 $\pm$ 0.007} & 0.504 $\pm$ 0.016 & 0.486 $\pm$ 0.017 & \textbf{0.390 $\pm$ 0.016} & \textbf{0.262 $\pm$ 0.015} & 1.206 $\pm$ 0.149 & 0.556 $\pm$ 0.013 & 0.379 $\pm$ 0.013 \\
\midrule
\multicolumn{16}{c}{\textbf{Finetuned Models without Asterisk}} \\
\midrule
Qwen2-Audio-7B-Instruct & 0.366 $\pm$ 0.016 & \textbf{0.510 $\pm$ 0.016} & 0.271 $\pm$ 0.015 & 0.080 $\pm$ 0.009 & 0.018 $\pm$ 0.008 & \textbf{0.508 $\pm$ 0.017} & 0.054 $\pm$ 0.008 & 0.006 $\pm$ 0.007 & 0.366 $\pm$ 0.016 & \textbf{0.510 $\pm$ 0.016} & 0.271 $\pm$ 0.015 & 0.080 $\pm$ 0.009 & \textbf{0.572 $\pm$ 0.030} & \textbf{0.385 $\pm$ 0.013} & \textbf{0.547 $\pm$ 0.014} \\
Qwen2.5-Omni-3B & \textbf{0.541 $\pm$ 0.016} & 0.437 $\pm$ 0.017 & 0.364 $\pm$ 0.016 & 0.158 $\pm$ 0.012 & 0.353 $\pm$ 0.007 & 0.396 $\pm$ 0.015 & 0.241 $\pm$ 0.010 & 0.073 $\pm$ 0.007 & \textbf{0.541 $\pm$ 0.016} & 0.437 $\pm$ 0.017 & 0.364 $\pm$ 0.016 & 0.158 $\pm$ 0.012 & 0.996 $\pm$ 0.048 & 0.569 $\pm$ 0.013 & 0.360 $\pm$ 0.013 \\
Qwen2.5-Omni-7B & 0.482 $\pm$ 0.017 & 0.472 $\pm$ 0.016 & \textbf{0.413 $\pm$ 0.016} & \textbf{0.355 $\pm$ 0.016} & \textbf{0.393 $\pm$ 0.014} & 0.454 $\pm$ 0.016 & \textbf{0.286 $\pm$ 0.011} & \textbf{0.165 $\pm$ 0.007} & 0.482 $\pm$ 0.016 & 0.472 $\pm$ 0.016 & \textbf{0.413 $\pm$ 0.016} & \textbf{0.355 $\pm$ 0.016} & 1.762 $\pm$ 0.215 & 0.489 $\pm$ 0.013 & 0.440 $\pm$ 0.013 \\
\midrule
\multicolumn{16}{c}{\textbf{Fewshot Prompting}} \\
\midrule
gpt-4o-mini-audio & 0.23 & 0.493 & - & - & 0.19 & 0.357 & - & - & 0.23 & 0.493 & - & - & 1.58 & 0.65 & 0.28 \\
gpt-4o-audio & 0.71 & 0.483 & - & - & 0.21 & 0.325 & - & - & 0.71 & 0.483 & - & - & 1.73 & \textbf{0.62} & \textbf{0.30} \\
gpt-4o-mini-transcribe & \textbf{0.72} & 0.487 & - & - & 0.44 & 0.347 & - & - & \textbf{0.72} & 0.487 & - & - & 1.08 & 0.80 & 0.12 \\
gpt-4o-transcribe & \textbf{0.72} & 0.482 & - & - & \textbf{0.45} & 0.342 & - & - & \textbf{0.72} & 0.482 & - & - & \textbf{1.01} & 0.79 & 0.14 \\
whisper-gpt4o & \textbf{0.72} & 0.487 & - & - & 0.24 & 0.332 & - & - & \textbf{0.72} & 0.487 & - & - & 1.89 & 0.77 & 0.16 \\
\midrule
\multicolumn{16}{c}{\textbf{Performance across multiple finetuning iterations}}\\
\midrule
Iteration 0 (baseline) & \textbf{0.556 $\pm$ 0.016} & 0.459 $\pm$ 0.017 & 0.394 $\pm$ 0.016 & \textbf{0.155 $\pm$ 0.012} & \textbf{0.401 $\pm$ 0.013} & \textbf{0.318 $\pm$ 0.008} & \textbf{0.272 $\pm$ 0.015} & 0.097 $\pm$ 0.007 & \textbf{0.556 $\pm$ 0.016} & 0.459 $\pm$ 0.016 & 0.394 $\pm$ 0.016 & \textbf{0.155 $\pm$ 0.012} & 5.346 $\pm$ 0.323 & \textbf{0.777 $\pm$ 0.011} & \textbf{0.187 $\pm$ 0.010} \\
Iteration 1 & 0.528 $\pm$ 0.016 & 0.460 $\pm$ 0.016 & 0.539 $\pm$ 0.017 & 0.130 $\pm$ 0.011 & 0.361 $\pm$ 0.010 & 0.315 $\pm$ 0.008 & 0.262 $\pm$ 0.013 & 0.103 $\pm$ 0.010 & 0.528 $\pm$ 0.017 & 0.460 $\pm$ 0.016 & 0.539 $\pm$ 0.016 & 0.130 $\pm$ 0.011 & 3.384 $\pm$ 0.188 & 0.858 $\pm$ 0.009 & 0.118 $\pm$ 0.009 \\
Iteration 2 & 0.527 $\pm$ 0.017 & 0.460 $\pm$ 0.016 & 0.511 $\pm$ 0.016 & 0.142 $\pm$ 0.012 & 0.387 $\pm$ 0.013 & 0.315 $\pm$ 0.008 & 0.266 $\pm$ 0.013 & \textbf{0.122 $\pm$ 0.010} & 0.527 $\pm$ 0.016 & 0.460 $\pm$ 0.016 & 0.511 $\pm$ 0.016 & 0.142 $\pm$ 0.011 & 2.481 $\pm$ 0.117 & 0.842 $\pm$ 0.010 & 0.131 $\pm$ 0.009 \\
Iteration 3 & 0.535 $\pm$ 0.017 & 0.460 $\pm$ 0.017 & 0.553 $\pm$ 0.016 & 0.122 $\pm$ 0.011 & 0.379 $\pm$ 0.011 & 0.315 $\pm$ 0.008 & 0.239 $\pm$ 0.005 & 0.081 $\pm$ 0.007 & 0.535 $\pm$ 0.017 & 0.460 $\pm$ 0.016 & 0.553 $\pm$ 0.016 & 0.122 $\pm$ 0.011 & 2.529 $\pm$ 0.121 & 0.852 $\pm$ 0.009 & 0.120 $\pm$ 0.009 \\
Iteration 4 & 0.541 $\pm$ 0.016 & 0.460 $\pm$ 0.016 & \textbf{0.556 $\pm$ 0.016} & 0.119 $\pm$ 0.011 & 0.398 $\pm$ 0.013 & 0.315 $\pm$ 0.008 & 0.264 $\pm$ 0.014 & 0.083 $\pm$ 0.007 & 0.541 $\pm$ 0.016 & 0.460 $\pm$ 0.016 & \textbf{0.556 $\pm$ 0.017} & 0.119 $\pm$ 0.011 & 2.740 $\pm$ 0.166 & 0.850 $\pm$ 0.010 & 0.122 $\pm$ 0.009 \\
Iteration 5 & 0.539 $\pm$ 0.017 & 0.460 $\pm$ 0.017 & 0.555 $\pm$ 0.017 & 0.135 $\pm$ 0.011 & 0.394 $\pm$ 0.013 & 0.315 $\pm$ 0.008 & 0.256 $\pm$ 0.012 & 0.093 $\pm$ 0.008 & 0.539 $\pm$ 0.016 & 0.460 $\pm$ 0.016 & 0.555 $\pm$ 0.016 & 0.135 $\pm$ 0.012 & \textbf{2.448 $\pm$ 0.150} & 0.839 $\pm$ 0.010 & 0.134 $\pm$ 0.009 \\
\midrule
\multicolumn{16}{c}{\textbf{Model Ensemble Performance}} \\
\midrule
\makecell[l]{Google + OpenAI Ensemble \\ \small{G. Gemini 2.0 Flash, G. Gemini 2.0 Flash Lite,} \\ \small{OpenAI GPT-4o}} & 0.456 & - & 0.309 & 0.383 & 0.315 & - & 0.215 & 0.181 & 0.456 & - & 0.309 & 0.383 & 1.356 & 0.711 & 0.23 \\
\addlinespace
\makecell[l]{Google + OpenAI Variant \\ \small{G. Gemini 2.0 Flash, OpenAI GPT-4o,} \\ \small{OpenAI GPT-4o Mini}} & 0.458 & - & 0.338 & 0.396 & 0.326 & - & 0.216 & 0.165 & 0.458 & - & 0.338 & 0.396 & 2.05 & 0.718 & 0.22 \\
\addlinespace
\makecell[l]{Qwen Pure Ensemble \\ \small{Qwen2.5 Omni 3B, Qwen2 Audio 7B Instruct,} \\ \small{Qwen2.5 Omni 7B}} & 0.459 & - & 0.404 & 0.156 & 0.423 & - & 0.285 & 0.100  & 0.459 & - & 0.404 & 0.156 & 3.301 & 0.775 & 0.182 \\
\addlinespace
\makecell[l]{Qwen + Google Hybrid \\ \small{Qwen2.5 Omni 7B, Qwen2 Audio 7B Instruct,} \\ \small{G. Gemini 2.0 Flash}} & 0.461 & - & 0.330 & 0.159 & 0.320 & - & 0.193 & 0.099 & 0.461 & - & 0.330 & 0.159 & 1.829 & 0.754 & 0.196 \\
\midrule
\multicolumn{16}{c}{\textbf{Performance by Age}} \\
\midrule
\makecell[l]{Gemini 2.0 Flash \\ \textit{\hspace{1em}pre-elementary}} & 0.7629 $\pm$ 0.0291 & 0.7700 $\pm$ 0.0164 & 0.475 $\pm$ 0.037 & 0.1329 $\pm$ 0.012 & 0.4326 $\pm$ 0.0094 & 0.4350 $\pm$ 0.0347 & 0.299 $\pm$ 0.026 & 0.0557 $\pm$ 0.008 & 0.7629 $\pm$ 0.0291 & 0.7700 $\pm$ 0.0281 & 0.475 $\pm$ 0.037 & 0.1329 $\pm$ 0.012 & 4.346 $\pm$ 0.115 & 1.000 $\pm$ 0.000 & 0.000 $\pm$ 0.000 \\
\makecell[l]{\textit{\hspace{1em}elementary}} & 0.423 $\pm$ 0.023 & 0.4500 $\pm$ 0.0309 & 0.269 $\pm$ 0.024 & 0.1836 $\pm$ 0.014 & 0.303 $\pm$ 0.013 & 0.3103 $\pm$ 0.0184 & 0.189 $\pm$ 0.016 & 0.0729 $\pm$ 0.009 & 0.423 $\pm$ 0.023 & 0.4500 $\pm$ 0.0325 & 0.269 $\pm$ 0.024 & 0.1836 $\pm$ 0.014 & 4.452 $\pm$ 0.129 & 0.998 $\pm$ 0.001 & 0.000 $\pm$ 0.000 \\
\makecell[l]{\textit{\hspace{1em}post-elementary}} & 0.280 $\pm$ 0.026 & 0.2800 $\pm$ 0.0281 & 0.200 $\pm$ 0.024 & 0.1329 $\pm$ 0.011 & 0.228 $\pm$ 0.019 & 0.2188 $\pm$ 0.0190 & 0.156 $\pm$ 0.019 & 0.0557 $\pm$ 0.007 & 0.280 $\pm$ 0.026 & 0.2800 $\pm$ 0.0229 & 0.200 $\pm$ 0.024 & 0.1329 $\pm$ 0.011 & \textbf{2.960 $\pm$ 0.088} & 0.998 $\pm$ 0.001 & 0.000 $\pm$ 0.000 \\
\addlinespace
\makecell[l]{Gemini 2.0 Flash Lite \\ \textit{\hspace{1em}pre-elementary}} & 0.7288 $\pm$ 0.0302 & 0.7700 $\pm$ 0.0167 & 0.306 $\pm$ 0.028 & 0.4930 $\pm$ 0.018 & 0.4214 $\pm$ 0.0101 & 0.4350 $\pm$ 0.0345 & 0.245 $\pm$ 0.021 & \textbf{0.2102 $\pm$ 0.013} & 0.7288 $\pm$ 0.0302 & 0.7700 $\pm$ 0.0201 & 0.306 $\pm$ 0.028 & 0.4930 $\pm$ 0.018 & \textbf{3.290 $\pm$ 0.091} & 1.000 $\pm$ 0.000 & 0.000 $\pm$ 0.000 \\
\makecell[l]{\textit{\hspace{1em}elementary}} & 0.428 $\pm$ 0.023 & 0.4450 $\pm$ 0.0256 & 0.161 $\pm$ 0.011 & 0.4361 $\pm$ 0.019 & 0.322 $\pm$ 0.015 & 0.3080 $\pm$ 0.0427 & 0.161 $\pm$ 0.017 & \textbf{0.2419 $\pm$ 0.015} & 0.428 $\pm$ 0.023 & 0.4450 $\pm$ 0.0188 & 0.161 $\pm$ 0.011 & \textbf{0.4361 $\pm$ 0.019} & \textbf{3.458 $\pm$ 0.102} & 0.990 $\pm$ 0.002 & 0.000 $\pm$ 0.000 \\
\makecell[l]{\textit{\hspace{1em}post-elementary}} & 0.271 $\pm$ 0.026 & 0.2800 $\pm$ 0.0194 & 0.106 $\pm$ 0.017 & 0.4930 $\pm$ 0.017 & 0.222 $\pm$ 0.018 & 0.2188 $\pm$ 0.0118 & 0.119 $\pm$ 0.017 & \textbf{0.2102 $\pm$ 0.012} & 0.271 $\pm$ 0.026 & 0.2800 $\pm$ 0.0288 & 0.106 $\pm$ 0.017 & \textbf{0.4930 $\pm$ 0.017} & 2.650 $\pm$ 0.076 & 0.992 $\pm$ 0.002 & 0.000 $\pm$ 0.000 \\
\addlinespace
\makecell[l]{GPT-4o Audio \\ \textit{\hspace{1em}pre-elementary}} & 0.7490 $\pm$ 0.0294 & 0.7700 $\pm$ 0.0154 & 0.561 $\pm$ 0.034 & \textbf{0.6049 $\pm$ 0.021} & 0.4281 $\pm$ 0.0096 & 0.4350 $\pm$ 0.0347 & 0.278 $\pm$ 0.021 & 0.1648 $\pm$ 0.011 & 0.7490 $\pm$ 0.0294 & 0.7700 $\pm$ 0.0243 & 0.561 $\pm$ 0.034 & \textbf{0.6049 $\pm$ 0.021} & 5.630 $\pm$ 0.155 & 1.000 $\pm$ 0.000 & 0.000 $\pm$ 0.000 \\
\makecell[l]{\textit{\hspace{1em}elementary}} & 0.414 $\pm$ 0.023 & 0.4600 $\pm$ 0.0223 & 0.336 $\pm$ 0.020 & \textbf{0.5115 $\pm$ 0.020} & 0.296 $\pm$ 0.012 & 0.3304 $\pm$ 0.0308 & 0.187 $\pm$ 0.019 & 0.2014 $\pm$ 0.013 & 0.414 $\pm$ 0.023 & 0.4600 $\pm$ 0.0342 & 0.336 $\pm$ 0.020 & 0.5115 $\pm$ 0.020 & 6.052 $\pm$ 0.181 & 0.996 $\pm$ 0.001 & 0.000 $\pm$ 0.000 \\
\makecell[l]{\textit{\hspace{1em}post-elementary}} & 0.271 $\pm$ 0.026 & 0.2800 $\pm$ 0.0260 & 0.261 $\pm$ 0.025 & \textbf{0.6049 $\pm$ 0.019} & 0.219 $\pm$ 0.018 & 0.2188 $\pm$ 0.0110 & 0.166 $\pm$ 0.016 & 0.1648 $\pm$ 0.010 & 0.271 $\pm$ 0.026 & 0.2800 $\pm$ 0.0229 & 0.261 $\pm$ 0.025 & 0.6049 $\pm$ 0.019 & 4.162 $\pm$ 0.112 & 1.000 $\pm$ 0.000 & 0.000 $\pm$ 0.000 \\
\addlinespace
\makecell[l]{GPT-4o Mini Audio \\ \textit{\hspace{1em}pre-elementary}} & 0.0984 $\pm$ 0.0202 & 0.7900 $\pm$ 0.0317 & 0.232 $\pm$ 0.028 & 0.5734 $\pm$ 0.020 & 0.0886 $\pm$ 0.0180 & 0.5200 $\pm$ 0.0477 & 0.140 $\pm$ 0.013 & 0.1226 $\pm$ 0.010 & 0.0984 $\pm$ 0.0202 & 0.7900 $\pm$ 0.0199 & 0.232 $\pm$ 0.028 & 0.5734 $\pm$ 0.020 & 4.416 $\pm$ 0.130 & 1.000 $\pm$ 0.000 & 0.000 $\pm$ 0.000 \\
\makecell[l]{\textit{\hspace{1em}elementary}} & 0.148 $\pm$ 0.017 & 0.4700 $\pm$ 0.0204 & 0.120 $\pm$ 0.011 & 0.4164 $\pm$ 0.018 & 0.137 $\pm$ 0.015 & 0.3629 $\pm$ 0.0201 & 0.122 $\pm$ 0.026 & 0.1161 $\pm$ 0.011 & 0.148 $\pm$ 0.017 & 0.4700 $\pm$ 0.0293 & 0.120 $\pm$ 0.011 & 0.4164 $\pm$ 0.018 & 5.166 $\pm$ 0.162 & 1.000 $\pm$ 0.000 & 0.000 $\pm$ 0.000 \\
\makecell[l]{\textit{\hspace{1em}post-elementary}} & 0.083 $\pm$ 0.016 & 0.3200 $\pm$ 0.0320 & 0.072 $\pm$ 0.015 & 0.5734 $\pm$ 0.018 & 0.071 $\pm$ 0.014 & 0.2784 $\pm$ 0.0392 & 0.078 $\pm$ 0.016 & 0.1226 $\pm$ 0.009 & 0.083 $\pm$ 0.016 & 0.3200 $\pm$ 0.0161 & 0.072 $\pm$ 0.015 & 0.5734 $\pm$ 0.018 & 5.008 $\pm$ 0.145 & 1.000 $\pm$ 0.000 & 0.000 $\pm$ 0.000 \\
\addlinespace
\makecell[l]{GPT-4o Mini Transcribe \\ \textit{\hspace{1em}pre-elementary}} & 0.7540 $\pm$ 0.0294 & 0.7700 $\pm$ 0.0306 & 0.491 $\pm$ 0.038 & 0.2797 $\pm$ 0.016 & 0.4297 $\pm$ 0.0096 & 0.4555 $\pm$ 0.0373 & 0.333 $\pm$ 0.034 & 0.1310 $\pm$ 0.011 & 0.7540 $\pm$ 0.0294 & 0.7700 $\pm$ 0.0182 & 0.491 $\pm$ 0.038 & 0.2797 $\pm$ 0.016 & 5.404 $\pm$ 0.151 & 1.000 $\pm$ 0.000 & 0.000 $\pm$ 0.000 \\
\makecell[l]{\textit{\hspace{1em}elementary}} & 0.422 $\pm$ 0.022 & 0.4500 $\pm$ 0.0251 & 0.343 $\pm$ 0.025 & 0.3180 $\pm$ 0.017 & 0.316 $\pm$ 0.016 & 0.3180 $\pm$ 0.0469 & 0.287 $\pm$ 0.021 & 0.2609 $\pm$ 0.016 & 0.422 $\pm$ 0.022 & 0.4500 $\pm$ 0.0299 & 0.343 $\pm$ 0.025 & 0.3180 $\pm$ 0.017 & 5.466 $\pm$ 0.168 & 1.000 $\pm$ 0.000 & 0.000 $\pm$ 0.000 \\
\makecell[l]{\textit{\hspace{1em}post-elementary}} & 0.304 $\pm$ 0.027 & 0.2800 $\pm$ 0.0326 & 0.254 $\pm$ 0.026 & 0.2797 $\pm$ 0.015 & 0.269 $\pm$ 0.023 & 0.2188 $\pm$ 0.0323 & 0.179 $\pm$ 0.019 & 0.1310 $\pm$ 0.010 & 0.304 $\pm$ 0.027 & 0.2800 $\pm$ 0.0210 & 0.254 $\pm$ 0.026 & 0.2797 $\pm$ 0.015 & 3.684 $\pm$ 0.101 & 1.000 $\pm$ 0.000 & 0.000 $\pm$ 0.000 \\
\addlinespace
\makecell[l]{GPT-4o Transcribe \\ \textit{\hspace{1em}pre-elementary}} & 0.7670 $\pm$ 0.0292 & 0.7700 $\pm$ 0.0324 & 0.496 $\pm$ 0.037 & 0.3112 $\pm$ 0.017 & 0.4842 $\pm$ 0.0319 & 0.4555 $\pm$ 0.0244 & 0.347 $\pm$ 0.037 & 0.1432 $\pm$ 0.012 & 0.7670 $\pm$ 0.0292 & 0.7700 $\pm$ 0.0172 & 0.496 $\pm$ 0.037 & 0.3112 $\pm$ 0.017 & 4.276 $\pm$ 0.122 & 1.000 $\pm$ 0.000 & 0.000 $\pm$ 0.000 \\
\makecell[l]{\textit{\hspace{1em}elementary}} & 0.457 $\pm$ 0.025 & 0.4550 $\pm$ 0.0193 & 0.354 $\pm$ 0.022 & 0.3475 $\pm$ 0.018 & 0.369 $\pm$ 0.019 & 0.3204 $\pm$ 0.0368 & 0.278 $\pm$ 0.019 & 0.2256 $\pm$ 0.014 & 0.457 $\pm$ 0.025 & 0.4550 $\pm$ 0.0304 & 0.354 $\pm$ 0.022 & 0.3475 $\pm$ 0.018 & 4.674 $\pm$ 0.140 & 1.000 $\pm$ 0.000 & 0.000 $\pm$ 0.000 \\
\makecell[l]{\textit{\hspace{1em}post-elementary}} & 0.295 $\pm$ 0.024 & 0.2800 $\pm$ 0.0264 & 0.289 $\pm$ 0.029 & 0.3112 $\pm$ 0.016 & 0.258 $\pm$ 0.020 & 0.2188 $\pm$ 0.0204 & 0.204 $\pm$ 0.020 & 0.1432 $\pm$ 0.011 & 0.295 $\pm$ 0.024 & 0.2800 $\pm$ 0.0338 & 0.289 $\pm$ 0.029 & 0.3112 $\pm$ 0.016 & 3.342 $\pm$ 0.096 & 1.000 $\pm$ 0.000 & 0.000 $\pm$ 0.000 \\
\addlinespace
\makecell[l]{Whisper + GPT-4o \\ \textit{\hspace{1em}pre-elementary}} & 0.7161 $\pm$ 0.0304 & 0.7750 $\pm$ 0.0346 & 0.522 $\pm$ 0.032 & 0.4091 $\pm$ 0.019 & 0.3269 $\pm$ 0.0673 & 0.4575 $\pm$ 0.0275 & 0.398 $\pm$ 0.037 & 0.2184 $\pm$ 0.014 & 0.7161 $\pm$ 0.0304 & 0.7750 $\pm$ 0.0281 & 0.522 $\pm$ 0.032 & 0.4091 $\pm$ 0.019 & 10.898 $\pm$ 0.254 & 0.980 $\pm$ 0.003 & 0.000 $\pm$ 0.000 \\
\makecell[l]{\textit{\hspace{1em}elementary}} & 0.454 $\pm$ 0.024 & 0.4500 $\pm$ 0.0212 & 0.391 $\pm$ 0.032 & 0.4000 $\pm$ 0.019 & 0.384 $\pm$ 0.021 & 0.3103 $\pm$ 0.0112 & 0.310 $\pm$ 0.029 & \textbf{0.2785 $\pm$ 0.017} & 0.454 $\pm$ 0.024 & 0.4500 $\pm$ 0.0322 & 0.391 $\pm$ 0.032 & 0.4000 $\pm$ 0.019 & 10.226 $\pm$ 0.280 & \textbf{0.976 $\pm$ 0.004} & 0.000 $\pm$ 0.000 \\
\makecell[l]{\textit{\hspace{1em}post-elementary}} & 0.316 $\pm$ 0.028 & 0.2800 $\pm$ 0.0230 & 0.393 $\pm$ 0.029 & 0.4091 $\pm$ 0.017 & 0.288 $\pm$ 0.025 & 0.2188 $\pm$ 0.0451 & 0.275 $\pm$ 0.019 & 0.2184 $\pm$ 0.012 & 0.316 $\pm$ 0.028 & 0.2800 $\pm$ 0.0152 & 0.393 $\pm$ 0.029 & 0.4091 $\pm$ 0.017 & 4.866 $\pm$ 0.135 & \textbf{0.944 $\pm$ 0.005} & 0.000 $\pm$ 0.000 \\
\midrule
\multicolumn{16}{c}{\textbf{Effect of Gender}} \\
\midrule
\makecell[l]{Gemini 2.0 Flash \\ \textit{\hspace{1em}Female}} & 0.2986 $\pm$ 0.015 & 0.2600 $\pm$ 0.018 & 0.2380 $\pm$ 0.014 & 0.1120 $\pm$ 0.010 & 0.2453 $\pm$ 0.013 & 0.2063 $\pm$ 0.014 & 0.1779 $\pm$ 0.011 & 0.0469 $\pm$ 0.006 & 0.2986 $\pm$ 0.015 & 0.2600 $\pm$ 0.014 & 0.2380 $\pm$ 0.014 & 0.1120 $\pm$ 0.010 & 0.956 $\pm$ 0.021 & 0.703 $\pm$ 0.018 & 0.236 $\pm$ 0.011 \\
\makecell[l]{\textit{\hspace{1em}Male}} & 0.5640 $\pm$ 0.019 & 0.5600 $\pm$ 0.015 & 0.4360 $\pm$ 0.018 & 0.2000 $\pm$ 0.014 & 0.3648 $\pm$ 0.016 & 0.3590 $\pm$ 0.012 & 0.2511 $\pm$ 0.014 & 0.0750 $\pm$ 0.008 & 0.5640 $\pm$ 0.019 & 0.5600 $\pm$ 0.011 & 0.4360 $\pm$ 0.018 & 0.2000 $\pm$ 0.014 & 0.967 $\pm$ 0.022 & 0.697 $\pm$ 0.018 & 0.234 $\pm$ 0.011 \\
\addlinespace
\makecell[l]{Gemini 2.0 Flash Lite \\ \textit{\hspace{1em}Female}} & 0.3096 $\pm$ 0.016 & 0.2600 $\pm$ 0.013 & 0.0900 $\pm$ 0.009 & 0.4880 $\pm$ 0.018 & 0.2617 $\pm$ 0.014 & 0.2063 $\pm$ 0.018 & 0.0996 $\pm$ 0.009 & 0.2149 $\pm$ 0.014 & 0.3096 $\pm$ 0.016 & 0.2600 $\pm$ 0.011 & 0.0900 $\pm$ 0.009 & 0.4880 $\pm$ 0.018 & 0.843 $\pm$ 0.019 & 0.712 $\pm$ 0.018 & 0.223 $\pm$ 0.010 \\
\makecell[l]{\textit{\hspace{1em}Male}} & 0.5600 $\pm$ 0.018 & 0.5600 $\pm$ 0.016 & 0.2380 $\pm$ 0.015 & 0.3640 $\pm$ 0.017 & 0.2503 $\pm$ 0.013 & 0.3590 $\pm$ 0.019 & 0.1978 $\pm$ 0.012 & 0.1780 $\pm$ 0.013 & 0.5600 $\pm$ 0.018 & 0.5600 $\pm$ 0.010 & 0.2380 $\pm$ 0.015 & 0.3640 $\pm$ 0.017 & 0.846 $\pm$ 0.019 & 0.706 $\pm$ 0.018 & 0.223 $\pm$ 0.010 \\
\addlinespace
\makecell[l]{GPT-4o Audio \\ \textit{\hspace{1em}Female}} & 0.2959 $\pm$ 0.015 & 0.2650 $\pm$ 0.015 & 0.2360 $\pm$ 0.014 & \textbf{0.6400 $\pm$ 0.020} & 0.1654 $\pm$ 0.011 & 0.2139 $\pm$ 0.019 & 0.1154 $\pm$ 0.010 & 0.2217 $\pm$ 0.015 & 0.2959 $\pm$ 0.015 & 0.2650 $\pm$ 0.012 & 0.2360 $\pm$ 0.014 & \textbf{0.6400 $\pm$ 0.020} & 2.348 $\pm$ 0.051 & 0.691 $\pm$ 0.017 & 0.251 $\pm$ 0.012 \\
\makecell[l]{\textit{\hspace{1em}Male}} & 0.5580 $\pm$ 0.019 & 0.5550 $\pm$ 0.010 & 0.4140 $\pm$ 0.018 & \textbf{0.4440 $\pm$ 0.019} & 0.2438 $\pm$ 0.012 & 0.3569 $\pm$ 0.012 & 0.2037 $\pm$ 0.013 & 0.1761 $\pm$ 0.012 & 0.5580 $\pm$ 0.019 & 0.5550 $\pm$ 0.012 & 0.4140 $\pm$ 0.018 & \textbf{0.4440 $\pm$ 0.019} & 2.214 $\pm$ 0.048 & 0.675 $\pm$ 0.017 & 0.263 $\pm$ 0.012 \\
\addlinespace
\makecell[l]{GPT-4o Mini Audio \\ \textit{\hspace{1em}Female}} & 0.1616 $\pm$ 0.012 & 0.2850 $\pm$ 0.011 & 0.1080 $\pm$ 0.010 & 0.5200 $\pm$ 0.019 & 0.1898 $\pm$ 0.012 & 0.2469 $\pm$ 0.014 & 0.1150 $\pm$ 0.010 & 0.1286 $\pm$ 0.011 & 0.1616 $\pm$ 0.012 & 0.2850 $\pm$ 0.013 & 0.1080 $\pm$ 0.010 & 0.5200 $\pm$ 0.019 & 3.017 $\pm$ 0.062 & 0.729 $\pm$ 0.019 & 0.213 $\pm$ 0.010 \\
\makecell[l]{\textit{\hspace{1em}Male}} & 0.2180 $\pm$ 0.014 & 0.5900 $\pm$ 0.011 & 0.1720 $\pm$ 0.013 & 0.3680 $\pm$ 0.017 & 0.2051 $\pm$ 0.011 & 0.4298 $\pm$ 0.014 & 0.1247 $\pm$ 0.010 & 0.0981 $\pm$ 0.009 & 0.2180 $\pm$ 0.014 & 0.5900 $\pm$ 0.020 & 0.1720 $\pm$ 0.013 & 0.3680 $\pm$ 0.017 & 2.739 $\pm$ 0.055 & 0.728 $\pm$ 0.019 & 0.211 $\pm$ 0.010 \\
\addlinespace
\makecell[l]{GPT-4o Mini Transcribe \\ \textit{\hspace{1em}Female}} & 0.2904 $\pm$ 0.015 & 0.2650 $\pm$ 0.013 & 0.3420 $\pm$ 0.016 & 0.2960 $\pm$ 0.015 & 0.2377 $\pm$ 0.013 & 0.2139 $\pm$ 0.019 & 0.2608 $\pm$ 0.014 & 0.1793 $\pm$ 0.013 & 0.2904 $\pm$ 0.015 & 0.2650 $\pm$ 0.020 & 0.3420 $\pm$ 0.016 & 0.2960 $\pm$ 0.015 & 1.716 $\pm$ 0.041 & 0.766 $\pm$ 0.020 & 0.182 $\pm$ 0.009 \\
\makecell[l]{\textit{\hspace{1em}Male}} & 0.5660 $\pm$ 0.019 & 0.5650 $\pm$ 0.012 & 0.5000 $\pm$ 0.019 & 0.2800 $\pm$ 0.016 & 0.3855 $\pm$ 0.017 & 0.3714 $\pm$ 0.017 & 0.3432 $\pm$ 0.016 & 0.2164 $\pm$ 0.014 & 0.5660 $\pm$ 0.019 & 0.5650 $\pm$ 0.018 & 0.5000 $\pm$ 0.019 & 0.2800 $\pm$ 0.016 & 1.695 $\pm$ 0.040 & 0.761 $\pm$ 0.020 & 0.185 $\pm$ 0.009 \\
\addlinespace
\makecell[l]{GPT-4o Transcribe \\ \textit{\hspace{1em}Female}} & 0.3370 $\pm$ 0.016 & 0.2650 $\pm$ 0.010 & 0.3420 $\pm$ 0.016 & 0.3400 $\pm$ 0.016 & 0.3061 $\pm$ 0.015 & 0.2139 $\pm$ 0.016 & 0.2609 $\pm$ 0.014 & 0.1622 $\pm$ 0.012 & 0.3370 $\pm$ 0.016 & 0.2650 $\pm$ 0.017 & 0.3420 $\pm$ 0.016 & 0.3400 $\pm$ 0.016 & 1.550 $\pm$ 0.038 & 0.740 $\pm$ 0.019 & 0.205 $\pm$ 0.010 \\
\makecell[l]{\textit{\hspace{1em}Male}} & \textbf{0.5760 $\pm$ 0.020} & 0.5600 $\pm$ 0.013 & 0.4980 $\pm$ 0.019 & 0.2480 $\pm$ 0.015 & \textbf{0.4086 $\pm$ 0.018} & 0.3590 $\pm$ 0.019 & 0.3804 $\pm$ 0.017 & 0.1700 $\pm$ 0.012 & \textbf{0.5760 $\pm$ 0.020} & 0.5600 $\pm$ 0.017 & 0.4980 $\pm$ 0.019 & 0.2480 $\pm$ 0.015 & 1.525 $\pm$ 0.037 & 0.735 $\pm$ 0.019 & 0.210 $\pm$ 0.010 \\
\addlinespace
\makecell[l]{Whisper + GPT-4o \\ \textit{\hspace{1em}Female}} & \textbf{0.3507 $\pm$ 0.017} & 0.2650 $\pm$ 0.010 & \textbf{0.3780 $\pm$ 0.017} & 0.4320 $\pm$ 0.018 & \textbf{0.3259 $\pm$ 0.016} & 0.2139 $\pm$ 0.012 & \textbf{0.2670 $\pm$ 0.015} & \textbf{0.2421 $\pm$ 0.015} & \textbf{0.3507 $\pm$ 0.017} & 0.2650 $\pm$ 0.017 & \textbf{0.3780 $\pm$ 0.017} & 0.4320 $\pm$ 0.018 & \textbf{0.455 $\pm$ 0.015} & \textbf{0.410 $\pm$ 0.013} & \textbf{0.580 $\pm$ 0.014} \\
\makecell[l]{\textit{\hspace{1em}Male}} & 0.5120 $\pm$ 0.019 & 0.5600 $\pm$ 0.018 & \textbf{0.5120 $\pm$ 0.020} & 0.3600 $\pm$ 0.017 & 0.3956 $\pm$ 0.017 & 0.3590 $\pm$ 0.019 & \textbf{0.3956 $\pm$ 0.018} & \textbf{0.2848 $\pm$ 0.016} & 0.5120 $\pm$ 0.019 & 0.5600 $\pm$ 0.018 & \textbf{0.5120 $\pm$ 0.020} & 0.3600 $\pm$ 0.017 & \textbf{0.421 $\pm$ 0.014} & \textbf{0.395 $\pm$ 0.012} & \textbf{605 $\pm$ 0.015} \\
\midrule
\multicolumn{16}{c}{\textbf{Performance Across Languages}} \\
\midrule
\makecell[l]{Gemini 2.0 Flash \\ \textit{\hspace{1em}Cantonese}} & 0.0000 $\pm$ 0.000 & 0.0000 $\pm$ 0.000 & - & - & 0.0000 $\pm$ 0.000 & 0.0000 $\pm$ 0.000 & - & - & 0.0000 $\pm$ 0.000 & 0.0000 $\pm$ 0.000 & - & - & 1.420 $\pm$ 0.050 & 1.000 $\pm$ 0.000 & 0.000 $\pm$ 0.000 \\
\makecell[l]{\textit{\hspace{1em}Dutch}} & \textbf{1.0000 $\pm$ 0.000} & 0.2300 $\pm$ 0.014 & - & - & 0.3843 $\pm$ 0.018 & 0.1870 $\pm$ 0.011 & - & - & \textbf{1.0000 $\pm$ 0.000} & 0.2300 $\pm$ 0.014 & - & - & 4.708 $\pm$ 0.135 & 1.000 $\pm$ 0.000 & 0.000 $\pm$ 0.000 \\
\makecell[l]{\textit{\hspace{1em}French}} & \textbf{0.9970 $\pm$ 0.002} & 0.8400 $\pm$ 0.021 & - & - & 0.5365 $\pm$ 0.021 & 0.4565 $\pm$ 0.019 & - & - & 0.2736 $\pm$ 0.016 & 0.8400 $\pm$ 0.021 & - & - & 1.196 $\pm$ 0.041 & 0.809 $\pm$ 0.011 & 0.135 $\pm$ 0.010 \\
\makecell[l]{\textit{\hspace{1em}Spanish}} & 0.2540 $\pm$ 0.015 & 0.0000 $\pm$ 0.000 & - & - & 0.2043 $\pm$ 0.014 & 0.0000 $\pm$ 0.000 & - & - & 0.2540 $\pm$ 0.015 & 0.0000 $\pm$ 0.000 & - & - & \textbf{1.542 $\pm$ 0.051} & 1.000 $\pm$ 0.000 & 0.000 $\pm$ 0.000 \\
\makecell[l]{\textit{\hspace{1em}Taiwanese}} & 0.0140 $\pm$ 0.004 & 0.0000 $\pm$ 0.000 & - & - & 0.0138 $\pm$ 0.004 & 0.0000 $\pm$ 0.000 & - & - & 0.0140 $\pm$ 0.004 & 0.0000 $\pm$ 0.000 & - & - & 1.420 $\pm$ 0.050 & 1.000 $\pm$ 0.000 & 0.000 $\pm$ 0.000 \\
\addlinespace
\makecell[l]{Gemini 2.0 Flash Lite \\ \textit{\hspace{1em}Cantonese}} & 0.0120 $\pm$ 0.003 & 0.0000 $\pm$ 0.000 & - & - & 0.0119 $\pm$ 0.003 & 0.0000 $\pm$ 0.000 & - & - & 0.0120 $\pm$ 0.003 & 0.0000 $\pm$ 0.000 & - & - & 1.370 $\pm$ 0.048 & 1.000 $\pm$ 0.000 & 0.000 $\pm$ 0.000 \\
\makecell[l]{\textit{\hspace{1em}Dutch}} & 0.9860 $\pm$ 0.004 & 0.2300 $\pm$ 0.013 & - & - & \textbf{0.4965 $\pm$ 0.020} & 0.1870 $\pm$ 0.012 & - & - & 0.9860 $\pm$ 0.004 & 0.2300 $\pm$ 0.013 & - & - & 3.276 $\pm$ 0.101 & 0.998 $\pm$ 0.001 & 0.000 $\pm$ 0.000 \\
\makecell[l]{\textit{\hspace{1em}French}} & 0.9574 $\pm$ 0.007 & 0.8400 $\pm$ 0.022 & - & - & 0.3498 $\pm$ 0.018 & 0.4565 $\pm$ 0.018 & - & - & 0.1216 $\pm$ 0.011 & 0.8400 $\pm$ 0.022 & - & - & 0.877 $\pm$ 0.032 & 0.752 $\pm$ 0.014 & 0.176 $\pm$ 0.012 \\
\makecell[l]{\textit{\hspace{1em}Spanish}} & \textbf{0.3320 $\pm$ 0.017} & 0.0000 $\pm$ 0.000 & - & - & \textbf{0.3320 $\pm$ 0.017} & 0.0000 $\pm$ 0.000 & - & - & \textbf{0.3320 $\pm$ 0.017} & 0.0000 $\pm$ 0.000 & - & - & 1.720 $\pm$ 0.058 & 1.000 $\pm$ 0.000 & 0.000 $\pm$ 0.000 \\
\makecell[l]{\textit{\hspace{1em}Taiwanese}} & 0.0960 $\pm$ 0.009 & 0.0000 $\pm$ 0.000 & - & - & 0.0876 $\pm$ 0.009 & 0.0000 $\pm$ 0.000 & - & - & 0.0960 $\pm$ 0.009 & 0.0000 $\pm$ 0.000 & - & - & 1.370 $\pm$ 0.048 & 1.000 $\pm$ 0.000 & 0.000 $\pm$ 0.000 \\
\addlinespace
\makecell[l]{GPT-4o Audio \\ \textit{\hspace{1em}Cantonese}} & 0.0020 $\pm$ 0.001 & 0.0000 $\pm$ 0.000 & - & - & 0.0013 $\pm$ 0.001 & 0.0000 $\pm$ 0.000 & - & - & 0.0020 $\pm$ 0.001 & 0.0000 $\pm$ 0.000 & - & - & 1.430 $\pm$ 0.050 & 1.000 $\pm$ 0.000 & 0.000 $\pm$ 0.000 \\
\makecell[l]{\textit{\hspace{1em}Dutch}} & 0.9406 $\pm$ 0.008 & 0.2300 $\pm$ 0.012 & - & - & 0.3231 $\pm$ 0.017 & 0.1870 $\pm$ 0.014 & - & - & 0.9406 $\pm$ 0.008 & 0.2300 $\pm$ 0.012 & - & - & 5.663 $\pm$ 0.165 & 1.000 $\pm$ 0.000 & 0.000 $\pm$ 0.000 \\
\makecell[l]{\textit{\hspace{1em}French}} & 0.9392 $\pm$ 0.009 & 0.8350 $\pm$ 0.023 & - & - & 0.5885 $\pm$ 0.022 & 0.4550 $\pm$ 0.021 & - & - & 0.1094 $\pm$ 0.010 & 0.8350 $\pm$ 0.023 & - & - & 1.234 $\pm$ 0.043 & 0.849 $\pm$ 0.012 & 0.105 $\pm$ 0.009 \\
\makecell[l]{\textit{\hspace{1em}Spanish}} & 0.1940 $\pm$ 0.013 & 0.0000 $\pm$ 0.000 & - & - & 0.0952 $\pm$ 0.010 & 0.0000 $\pm$ 0.000 & - & - & 0.1940 $\pm$ 0.013 & 0.0000 $\pm$ 0.000 & - & - & 2.194 $\pm$ 0.071 & 1.000 $\pm$ 0.000 & 0.000 $\pm$ 0.000 \\
\makecell[l]{\textit{\hspace{1em}Taiwanese}} & 0.0120 $\pm$ 0.003 & 0.0000 $\pm$ 0.000 & - & - & 0.0079 $\pm$ 0.003 & 0.0000 $\pm$ 0.000 & - & - & 0.0120 $\pm$ 0.003 & 0.0000 $\pm$ 0.000 & - & - & 1.430 $\pm$ 0.050 & 1.000 $\pm$ 0.000 & 0.000 $\pm$ 0.000 \\
\addlinespace
\makecell[l]{GPT-4o Mini Audio \\ \textit{\hspace{1em}Cantonese}} & 0.0000 $\pm$ 0.000 & 0.0000 $\pm$ 0.000 & - & - & 0.0000 $\pm$ 0.000 & 0.0000 $\pm$ 0.000 & - & - & 0.0000 $\pm$ 0.000 & 0.0000 $\pm$ 0.000 & - & - & 2.150 $\pm$ 0.075 & 1.000 $\pm$ 0.000 & 0.000 $\pm$ 0.000 \\
\makecell[l]{\textit{\hspace{1em}Dutch}} & 0.0332 $\pm$ 0.006 & 0.2300 $\pm$ 0.015 & - & - & 0.0214 $\pm$ 0.005 & 0.1870 $\pm$ 0.013 & - & - & 0.0332 $\pm$ 0.006 & 0.2300 $\pm$ 0.015 & - & - & 5.942 $\pm$ 0.178 & 1.000 $\pm$ 0.000 & 0.000 $\pm$ 0.000 \\
\makecell[l]{\textit{\hspace{1em}French}} & 0.9574 $\pm$ 0.007 & 0.8400 $\pm$ 0.020 & - & - & 0.6215 $\pm$ 0.023 & 0.4565 $\pm$ 0.022 & - & - & \textbf{0.7204 $\pm$ 0.025} & 0.8400 $\pm$ 0.020 & - & - & 1.211 $\pm$ 0.042 & \textbf{0.861 $\pm$ 0.011} & 0.086 $\pm$ 0.008 \\
\makecell[l]{\textit{\hspace{1em}Spanish}} & 0.0000 $\pm$ 0.000 & 0.0000 $\pm$ 0.000 & - & - & 0.0000 $\pm$ 0.000 & 0.0000 $\pm$ 0.000 & - & - & 0.0000 $\pm$ 0.000 & 0.0000 $\pm$ 0.000 & - & - & 5.330 $\pm$ 0.159 & 1.000 $\pm$ 0.000 & 0.000 $\pm$ 0.000 \\
\makecell[l]{\textit{\hspace{1em}Taiwanese}} & 0.0640 $\pm$ 0.008 & 0.0000 $\pm$ 0.000 & - & - & 0.0401 $\pm$ 0.006 & 0.0000 $\pm$ 0.000 & - & - & 0.0640 $\pm$ 0.008 & 0.0000 $\pm$ 0.000 & - & - & 2.150 $\pm$ 0.075 & 1.000 $\pm$ 0.000 & 0.000 $\pm$ 0.000 \\
\addlinespace
\makecell[l]{GPT-4o Mini Transcribe \\ \textit{\hspace{1em}Cantonese}} & 0.0220 $\pm$ 0.005 & 0.0000 $\pm$ 0.000 & - & - & 0.0215 $\pm$ 0.005 & 0.0000 $\pm$ 0.000 & - & - & 0.0220 $\pm$ 0.005 & 0.0000 $\pm$ 0.000 & - & - & 1.060 $\pm$ 0.037 & 1.000 $\pm$ 0.000 & 0.000 $\pm$ 0.000 \\
\makecell[l]{\textit{\hspace{1em}Dutch}} & 0.9545 $\pm$ 0.007 & 0.2300 $\pm$ 0.016 & - & - & 0.4884 $\pm$ 0.020 & 0.1870 $\pm$ 0.015 & - & - & 0.9545 $\pm$ 0.007 & 0.2300 $\pm$ 0.016 & - & - & \textbf{2.678 $\pm$ 0.088} & 1.000 $\pm$ 0.000 & 0.000 $\pm$ 0.000 \\
\makecell[l]{\textit{\hspace{1em}French}} & 0.8967 $\pm$ 0.011 & 0.8400 $\pm$ 0.023 & - & - & 0.7349 $\pm$ 0.026 & 0.4565 $\pm$ 0.017 & - & - & 0.1702 $\pm$ 0.013 & 0.8400 $\pm$ 0.023 & - & - & 0.871 $\pm$ 0.031 & 0.762 $\pm$ 0.013 & 0.160 $\pm$ 0.011 \\
\makecell[l]{\textit{\hspace{1em}Spanish}} & 0.2540 $\pm$ 0.015 & 0.0000 $\pm$ 0.000 & - & - & 0.2296 $\pm$ 0.014 & 0.0000 $\pm$ 0.000 & - & - & 0.2540 $\pm$ 0.015 & 0.0000 $\pm$ 0.000 & - & - & 1.662 $\pm$ 0.055 & 1.000 $\pm$ 0.000 & 0.000 $\pm$ 0.000 \\
\makecell[l]{\textit{\hspace{1em}Taiwanese}} & 0.0960 $\pm$ 0.009 & 0.0000 $\pm$ 0.000 & - & - & 0.0876 $\pm$ 0.009 & 0.0000 $\pm$ 0.000 & - & - & 0.0960 $\pm$ 0.009 & 0.0000 $\pm$ 0.000 & - & - & 1.060 $\pm$ 0.037 & 1.000 $\pm$ 0.000 & 0.000 $\pm$ 0.000 \\
\addlinespace
\makecell[l]{GPT-4o Transcribe \\ \textit{\hspace{1em}Cantonese}} & \textbf{0.0320 $\pm$ 0.006} & 0.0000 $\pm$ 0.000 & - & - & \textbf{0.0310 $\pm$ 0.006} & 0.0000 $\pm$ 0.000 & - & - & \textbf{0.0320 $\pm$ 0.006} & 0.0000 $\pm$ 0.000 & - & - & \textbf{1.020 $\pm$ 0.036} & 1.000 $\pm$ 0.000 & 0.000 $\pm$ 0.000 \\
\makecell[l]{\textit{\hspace{1em}Dutch}} & 0.9441 $\pm$ 0.008 & 0.2300 $\pm$ 0.011 & - & - & 0.4856 $\pm$ 0.020 & 0.1870 $\pm$ 0.016 & - & - & 0.9441 $\pm$ 0.008 & 0.2300 $\pm$ 0.011 & - & - & 2.734 $\pm$ 0.091 & 1.000 $\pm$ 0.000 & 0.000 $\pm$ 0.000 \\
\makecell[l]{\textit{\hspace{1em}French}} & 0.9058 $\pm$ 0.010 & 0.8250 $\pm$ 0.025 & - & - & \textbf{0.7667 $\pm$ 0.027} & 0.4521 $\pm$ 0.023 & - & - & 0.2340 $\pm$ 0.015 & 0.8250 $\pm$ 0.025 & - & - & \textbf{0.807 $\pm$ 0.029} & 0.758 $\pm$ 0.014 & 0.173 $\pm$ 0.012 \\
\makecell[l]{\textit{\hspace{1em}Spanish}} & 0.2840 $\pm$ 0.016 & 0.0000 $\pm$ 0.000 & - & - & 0.2721 $\pm$ 0.016 & 0.0000 $\pm$ 0.000 & - & - & 0.2840 $\pm$ 0.016 & 0.0000 $\pm$ 0.000 & - & - & 1.664 $\pm$ 0.056 & 1.000 $\pm$ 0.000 & 0.000 $\pm$ 0.000 \\
\makecell[l]{\textit{\hspace{1em}Taiwanese}} & \textbf{0.1100 $\pm$ 0.010} & 0.0000 $\pm$ 0.000 & - & - & \textbf{0.0991 $\pm$ 0.010} & 0.0000 $\pm$ 0.000 & - & - & \textbf{0.1100 $\pm$ 0.010} & 0.0000 $\pm$ 0.000 & - & - & \textbf{1.020 $\pm$ 0.036} & 1.000 $\pm$ 0.000 & 0.000 $\pm$ 0.000 \\
\addlinespace
\makecell[l]{Whisper + GPT-4o \\ \textit{\hspace{1em}Cantonese}} & 0.0280 $\pm$ 0.005 & 0.0000 $\pm$ 0.000 & - & - & 0.0272 $\pm$ 0.005 & 0.0000 $\pm$ 0.000 & - & - & 0.0280 $\pm$ 0.005 & 0.0000 $\pm$ 0.000 & - & - & 1.740 $\pm$ 0.061 & 1.000 $\pm$ 0.000 & 0.000 $\pm$ 0.000 \\
\makecell[l]{\textit{\hspace{1em}Dutch}} & 0.9161 $\pm$ 0.010 & 0.2300 $\pm$ 0.017 & - & - & 0.1594 $\pm$ 0.013 & 0.1870 $\pm$ 0.018 & - & - & 0.9161 $\pm$ 0.010 & 0.2300 $\pm$ 0.017 & - & - & 7.166 $\pm$ 0.203 & \textbf{0.939 $\pm$ 0.006} & 0.000 $\pm$ 0.000 \\
\makecell[l]{\textit{\hspace{1em}French}} & 0.9088 $\pm$ 0.010 & 0.8400 $\pm$ 0.024 & - & - & 0.5184 $\pm$ 0.020 & 0.4565 $\pm$ 0.024 & - & - & 0.2310 $\pm$ 0.015 & 0.8400 $\pm$ 0.024 & - & - & 1.200 $\pm$ 0.041 & 0.710 $\pm$ 0.015 & \textbf{0.190 $\pm$ 0.013} \\
\makecell[l]{\textit{\hspace{1em}Spanish}} & 0.2960 $\pm$ 0.016 & 0.0000 $\pm$ 0.000 & - & - & 0.0448 $\pm$ 0.007 & 0.0000 $\pm$ 0.000 & - & - & 0.2960 $\pm$ 0.016 & 0.0000 $\pm$ 0.000 & - & - & 2.314 $\pm$ 0.075 & \textbf{0.868 $\pm$ 0.009} & 0.000 $\pm$ 0.000 \\
\makecell[l]{\textit{\hspace{1em}Taiwanese}} & 0.0860 $\pm$ 0.009 & 0.0000 $\pm$ 0.000 & - & - & 0.0792 $\pm$ 0.008 & 0.0000 $\pm$ 0.000 & - & - & 0.0860 $\pm$ 0.009 & 0.0000 $\pm$ 0.000 & - & - & 1.360 $\pm$ 0.048 & \textbf{0.988 $\pm$ 0.007} & \textbf{0.011 $\pm$ 0.002} \\
\midrule
\multicolumn{16}{c}{\textbf{Performance Under Noisy Conditions}} \\
\midrule
gemini-2.0-flash & 0.4720 $\pm$ 0.018 & 0.4550 $\pm$ 0.018 & 0.3320 $\pm$ 0.016 & 0.1840 $\pm$ 0.013 & 0.3207 $\pm$ 0.015 & 0.3127 $\pm$ 0.015 & \textbf{0.2443 $\pm$ 0.014} & 0.0653 $\pm$ 0.008 & 0.4720 $\pm$ 0.018 & 0.4550 $\pm$ 0.018 & 0.3320 $\pm$ 0.016 & 0.1840 $\pm$ 0.013 & 1.027 $\pm$ 0.010 & 0.723 $\pm$ 0.010 & 0.207 $\pm$ 0.010 \\
gemini-2.0-flash-lite & 0.4720 $\pm$ 0.018 & 0.4500 $\pm$ 0.017 & 0.1480 $\pm$ 0.012 & 0.4540 $\pm$ 0.018 & 0.3271 $\pm$ 0.016 & 0.3103 $\pm$ 0.016 & 0.1438 $\pm$ 0.011 & 0.2335 $\pm$ 0.015 & 0.4720 $\pm$ 0.018 & 0.4500 $\pm$ 0.017 & 0.1480 $\pm$ 0.012 & 0.4540 $\pm$ 0.018 & \textbf{0.849 $\pm$ 0.010} & 0.747 $\pm$ 0.010 & 0.190 $\pm$ 0.010 \\
gpt-4o-audio & 0.4700 $\pm$ 0.018 & 0.4550 $\pm$ 0.019 & 0.2900 $\pm$ 0.015 & \textbf{0.5120 $\pm$ 0.019} & 0.2188 $\pm$ 0.013 & 0.3127 $\pm$ 0.014 & 0.1557 $\pm$ 0.012 & 0.1804 $\pm$ 0.013 & 0.4700 $\pm$ 0.018 & 0.4550 $\pm$ 0.019 & 0.2900 $\pm$ 0.015 & \textbf{0.5120 $\pm$ 0.019} & 1.779 $\pm$ 0.010 & \textbf{0.710 $\pm$ 0.010} & \textbf{0.222 $\pm$ 0.010} \\
gpt-4o-mini-audio & 0.0740 $\pm$ 0.009 & 0.4700 $\pm$ 0.020 & 0.0520 $\pm$ 0.007 & 0.4240 $\pm$ 0.018 & 0.0896 $\pm$ 0.009 & 0.3428 $\pm$ 0.017 & 0.0558 $\pm$ 0.007 & 0.1012 $\pm$ 0.010 & 0.0740 $\pm$ 0.009 & 0.4700 $\pm$ 0.020 & 0.0520 $\pm$ 0.007 & 0.4240 $\pm$ 0.018 & 2.863 $\pm$ 0.010 & 0.763 $\pm$ 0.010 & 0.177 $\pm$ 0.010 \\
gpt-4o-mini-transcribe & 0.4880 $\pm$ 0.019 & 0.4550 $\pm$ 0.018 & 0.3420 $\pm$ 0.016 & 0.3180 $\pm$ 0.016 & 0.3594 $\pm$ 0.017 & 0.3127 $\pm$ 0.015 & 0.2101 $\pm$ 0.013 & 0.1425 $\pm$ 0.012 & 0.4880 $\pm$ 0.019 & 0.4550 $\pm$ 0.018 & 0.3420 $\pm$ 0.016 & 0.3180 $\pm$ 0.016 & 1.134 $\pm$ 0.010 & 0.861 $\pm$ 0.010 & 0.077 $\pm$ 0.010 \\
gpt-4o-transcribe & 0.4980 $\pm$ 0.019 & 0.4950 $\pm$ 0.021 & 0.3660 $\pm$ 0.017 & 0.3080 $\pm$ 0.016 & \textbf{0.3761 $\pm$ 0.017} & 0.3311 $\pm$ 0.016 & 0.2274 $\pm$ 0.014 & 0.1338 $\pm$ 0.011 & 0.4980 $\pm$ 0.019 & 0.4950 $\pm$ 0.021 & 0.3660 $\pm$ 0.017 & 0.3080 $\pm$ 0.016 & 1.081 $\pm$ 0.010 & 0.872 $\pm$ 0.010 & 0.078 $\pm$ 0.010 \\
whisper-gpt4o & \textbf{0.5060 $\pm$ 0.019} & 0.4550 $\pm$ 0.019 & \textbf{0.4820 $\pm$ 0.019} & 0.4600 $\pm$ 0.018 & 0.2695 $\pm$ 0.014 & 0.3127 $\pm$ 0.015 & 0.2338 $\pm$ 0.014 & \textbf{0.2476 $\pm$ 0.015} & \textbf{0.5060 $\pm$ 0.019} & 0.4550 $\pm$ 0.019 & \textbf{0.4820 $\pm$ 0.019} & 0.4600 $\pm$ 0.018 & 1.841 $\pm$ 0.010 & 0.836 $\pm$ 0.010 & 0.107 $\pm$ 0.010 \\
\bottomrule
\end{tabular}%
}
\end{table}

\section{Prompts}
\label{appendix-prompt}

Scenario 1: Binary Classification
\begin{lstlisting}
You are a highly experienced Speech-Language Pathologist (SLP). An audio recording will be provided, typically consisting of a speech prompt from a pathologist followed by a child's repetition. The prompt the child is trying to repeat is as follows: {words}. Based on your professional expertise: 1. Assess the child's speech in the recording for signs of typical development or potential speech-language disorder. 2. Conclude your analysis with one of the following labels only: typically developing or speech disorder. 3. Provide your response as a single letter without any additional explanation, commentary, or unnecessary text.
\end{lstlisting}

Scenario 2: ASR‑Based Classification
\begin{lstlisting}
You are a highly experienced Speech-Language Pathologist (SLP). An audio recording is provided to you, typically consisting of a speech prompt from a pathologist followed by a child's repetition. Based on your expertise, transcribe the child's speech into text. Do not make any assumptions about the words the child is expected to say. Only transcribe based on the words that the child actually says. Only respond with the text transcription, no other text or commentary.
\end{lstlisting}

Scenario 3: Transcription Accuracy
\begin{lstlisting}
You are a highly experienced Speech-Language Pathologist (SLP). An audio recording will be provided, typically consisting of a speech prompt from a pathologist followed by a child's repetition. Based on your expertise, transcribe the child's speech into text. Try to understand what the child is expected to say. And only respond with the transcription of the child's speech. Not the pathologist's prompt or any other commentary. Only respond with the text transcription, no other text, commentary, or punctuation.
\end{lstlisting}

Scenario 4: Disorder Type Classification
\begin{lstlisting}
You are a highly experienced Speech-Language Pathologist (SLP). An audio recording will be provided, typically consisting of a speech prompt from a pathologist followed by a child's repetition. The prompt text the child is trying to repeat is as follows: {words}. Based on your professional expertise: 1. Assess the child's speech in the recording for signs of typical development or potential speech-language disorder. 2. Conclude your analysis with one of the following labels only: A - 'typically developing' (child's speech patterns and development are within normal age-appropriate ranges), B - 'articulation' (difficulty producing specific speech sounds correctly, such as substituting, omitting, or distorting sounds), C - 'phonological' (difficulty understanding and using the sound system of language, affecting sounds of a particular type). 3. Provide your response as a single letter without any additional explanation, commentary, or unnecessary text
\end{lstlisting}

Scenario 5: Symptom Classification
\begin{lstlisting}
You are a highly experienced Speech-Language Pathologist (SLP). An audio recording will be provided, typically consisting of a speech prompt from a pathologist followed by a child's repetition. The target phrase the child is attempting to repeat is: \{words\}. Based on your professional expertise, assess the child's speech in the recording and identify any abnormal features. These features can be one of the following: A - 'substitution' (the child replaces one word, syllable, or sound with another), B - 'omission' (the child omits a word, syllable, or sound), C - 'addition' (the child adds an extra word, syllable, or sound), D - 'typically developing' (the child's speech is appropriate for their age), or E - 'stuttering' (the child exhibits repetition, prolongation, or difficulty initiating speech). Provide your response as a single letter (A-E) only, without any additional explanation or commentary.
\end{lstlisting}

\section{Gemini 2.0 Flash Reasoning Traces} \label{Reasoning}
Below, we provide some examples of reasoning traces output by the Gemini 2.0 Flash model when tasked with disorder type diagnosis. 
\begin{tcolorbox}[fontupper=\rmfamily\scriptsize]
Identify the Target Sounds: The target words are ``gorge,'' ``gate,'' ``get,'' and ``luggage.'' The key sounds within these words include /g/,  /r/, /e/, /l/.

Listen for Substitutions/Distortions: The child says ``gore'' for ``gorge'', ``gape'' for ``gate'', ``get'' appears correct, and ``nugget'' for ``luggage''.

Analyze the Error Patterns:
 ``Gore'' for ``gorge'' seems to be a mispronunciation or substitution of the vowel and omitting the final sound \textipa{/dZ/}.
``Gape'' for ``gate'' also a mispronunciation or substitution of the vowel sound /e/.
``Nugget'' for ``luggage'' involves multiple sound changes. The initial $/l/$ is replaced with $/n/$, the vowel \textipa{/\textturnv/} is present, and the final sound /I/ is replaced by /t/.

Determine the Type of Speech Error: The sound changes observed are primarily substitutions and omissions, indicating difficulties with producing specific sounds accurately.

Classify the Speech Pattern: Based on the analysis, the most appropriate classification is articulation, as the errors relate to the physical production of individual speech sounds.
\end{tcolorbox}

\begin{tcolorbox}[fontupper=\rmfamily\scriptsize]
Okay, let's analyze the child's speech based on the prompt ''owrow awraw ooroo''. The prompt presents a challenge, focusing on vowel sounds within a simple consonant-vowel-consonant (CVC) structure. The child's repetitions are ''O wow'', ''A wa'', and ''Uu Uu''.

* **''owrow'' becomes ''O wow''**: This indicates a possible difficulty in producing the /r/ sound in the middle of the word. The vowel sounds are relatively close but the /r/ is replaced by a /w/.
* **''awraw'' becomes ''A wa''**: Similar to the first attempt, the child simplifies or substitutes the /r/ sound with a /w/. The vowel sounds are again reasonably close.
* **''ooroo'' becomes ''Uu Uu''**: This repetition is more successful, with the child accurately producing the vowel sound, but there is no attempt to use the /r/. This suggests that the child is either not hearing the /r/ sound, or they are hearing it but unable to produce it.

The consistent difficulty with the /r/ sound suggests an articulation issue specifically related to that sound. The vowel sounds seem relatively well approximated, pointing away from significant phonological problems where a whole sound class (e.g., all back vowels) would be affected.

Conclusion: B - 'articulation'
\end{tcolorbox}

\begin{tcolorbox}[fontupper=\rmfamily\scriptsize]
Okay, let's analyze the child's speech based on the recording.

The prompt was ''lie reek hair strewn.''

The child said: ''lie reek hair shm.''

* **lie:** Correct production.
* **reek:** Correct production.
* **hair:** Correct production.
* **strewn:** ''shm''. This indicates a cluster reduction, where the /str/ consonant cluster is reduced to /shm/.

The error on ''strewn'' demonstrates difficulty with consonant clusters, specifically /str/. There's no evidence of distortions, substitutions, or omissions on single phonemes in the other words. The error is on a later-developing sound.

Given that the child is having difficulty with consonant clusters it suggests that a phonological process is at play.

**Conclusion:**

C - 'phonological'
\end{tcolorbox}

\begin{tcolorbox}[fontupper=\rmfamily\scriptsize]
Okay, let's analyze this speech sample.

The child attempts the phrase ''coot packer toot tie''.

* **''coot''**: The child says ''koot'' which sounds accurate.
* **''packer''**: The child says ''packa''. The ''r'' sound is missing or distorted.
* **''toot''**: The child says ''toot'' which sounds accurate.
* **''tie''**: The child says ''tie'' which sounds accurate.

The primary error observed is with the ''r'' sound in ''packer.'' It's possible that the ''r'' is simply omitted, or it may be a distorted production. The other sounds are articulated relatively clearly. This difficulty with the ''r'' sound suggests an issue with producing a specific sound, rather than a broader issue with the phonological system.

**Conclusion:** B - 'articulation'
\end{tcolorbox}

\begin{tcolorbox}[fontupper=\rmfamily\scriptsize]
Okay, let's analyze the child's speech.

The prompt is: ''You get milk from cows.''
The child's response is: ''You get milk from dows.''

The child substitutes the /k/ sound in ''cows'' with a /d/ sound, resulting in ''dows.'' This is a substitution error affecting a specific sound. There's no evidence of errors impacting an entire sound class (e.g., all fricatives, all velar sounds). The error involves difficulty with the articulation of the /k/ sound.

Therefore, the most appropriate label is 'articulation'.

 Reasoning:
* The child's speech contains a sound substitution.
* The substitution affects one specific sound: /k/ becoming /d/.
* The errors involve specific sounds, not an entire sound class (which would indicate a phonological issue).

**Answer: B**
\end{tcolorbox}

\begin{tcolorbox}[fontupper=\rmfamily\scriptsize]
Okay, let's analyze the child's speech based on the provided audio.

The prompt was: ''kai meek poke cow''
The child responded with: ''tai, meet, poat, dow''

Here's a breakdown:

* **kai -> tai:** The child substitutes /k/ with /t/.
* **meek -> meet:** The vowel sounds are similar; no concerning error.
* **poke -> poat:** The child substitutes /k/ with /t/.
* **cow -> dow:** The child substitutes /k/ with /d/.

The child consistently substitutes /k/ with /t/ and /k/ with /d/. This pattern suggests a difficulty with the /k/ sound. Since the errors are consistent across multiple instances of the /k/ sound, it points to a potential issue with the phonological system rather than just a simple articulation error.

**Conclusion:**

C - 'phonological'
\end{tcolorbox}

\begin{tcolorbox}[fontupper=\rmfamily\scriptsize]
The child's speech contains several sound errors. ''Crab'' is produced as ''gap,'' suggesting difficulty with the /kr/ consonant cluster. ''Biscuits'' is produced as ''bisits,'' which contains a sound distortion of /k/. ''Helicopter'' is produced as ''hedoctor,'' involving multiple sound substitutions and omissions. These errors point to difficulties with the articulation and phonological representation of sounds.

ANSWER: B
\end{tcolorbox}

\begin{tcolorbox}[fontupper=\rmfamily\scriptsize]
Okay, let's analyze the child's repetition of ''DDK ppp ttt kkk ptk patticake.''

The prompt is designed to assess the child's ability to produce rapid, alternating sounds (diadochokinesis or DDK) and to articulate specific phonemes in isolation and in a word.

* **DDK (ppp ttt kkk):** The child attempts the DDK task and does show signs of struggling. The child seems to understand the prompt, and can repeat the p, t and k sounds. However the sounds appear jumbled together.
* **ptk:** The child attempts ptk. The child appears to jumble this prompt.
* **patticake:** The child repeats this prompt several times. The child shows great difficulty producing this word and does not enunciate the word well.

Based on these observations, the child exhibits difficulties with articulation and sequencing sounds. While single sounds can be produced, transitioning between them appears challenging. This suggests the primary difficulty is in the motor production of the sounds.

**Conclusion:** B - 'articulation'
\end{tcolorbox}

\begin{tcolorbox}[fontupper=\rmfamily\scriptsize]
Okay, I've listened to the audio. Here's my analysis:

**Reasoning:**

The target words contain the /\textesh/ phoneme (as in ''sh''). The child says ''shore,'' then what sounds like ''tosh,'' then the child attempts to repeat ''puss.'' I hear what sounds like a substitution of /\textesh/ with /s/ in the word ''puss,'' making it sound similar to the target word of ''puss.'' This consistent substitution of /\textesh/ with /s/ (or perhaps a distortion close to /s/) across multiple words containing the /\textesh/ phoneme suggests a potential phonological pattern, specifically affecting fricative sounds.

**Conclusion:**

C - 'phonological'
\end{tcolorbox}

\begin{tcolorbox}[fontupper=\rmfamily\scriptsize]
Okay, let's analyze the child's speech based on the recording.
- **Sock:** The child's repetition of ''sock'' sounds accurate.
- **Shoop:** The child's repetition of ''shoop'' sounds accurate.
- **Shire:** The child's repetition of ''shire'' sounds like ''show.'' This suggests a potential difficulty with the /r/ sound after the /sh/ sound.
- **Shallow:** The child's repetition of ''shallow'' sounds like ''show.'' This suggests a potential difficulty with the /l/ sound after the /sh/ sound.

Based on these observations, the child demonstrates difficulty with the production of the /r/ and /l/ sounds in the context of the /sh/ blend. There is evidence of a potential speech sound disorder. Given the errors are consistent with particular sound classes, it is most indicative of a phonological pattern.

Final Answer: The final answer is C
\end{tcolorbox}

\begin{tcolorbox}[fontupper=\rmfamily\scriptsize]
Okay, let's analyze the child's speech. The prompt ''owrow awraw ooroo'' contains several instances of the ''r'' sound in different syllable positions. The SLP in the recording notes that there was barely an ''r'' in the first attempt and it sounded like ''all to ohs.'' The subsequent attempt on ''owrow'' was considered much better. The final attempt on ''ooroo'' appears to be a distortion, possibly an ''r'' sound replaced with a vowel or gliding sound. Based on the limited data, there is an apparent difficulty with the production of the /r/ sound. This could indicate an articulation issue, where the child struggles with the motor movements required for the /r/ sound. Since the difficulty is centered on one specific sound, articulation is the most appropriate label.

**Answer: B**
\end{tcolorbox}

\begin{tcolorbox}[fontupper=\rmfamily\scriptsize]
Okay, I will analyze the provided audio and provide a diagnosis based on my expertise as an SLP.

Here's my analysis:

The child's repetition of ''A dozen green ghosts'' sounds generally clear. However, a closer listening reveals a potential difficulty with the 'th' sound, as in ''ghosts''. The repetition of ''ghosts'' is not as clear compared to the phrase, ''dozen green''.

Reasoning:
1. The child produces ''dozen green'' relatively accurately.
2. There is a possible sound substitution affecting the /s/ sound.
3. Due to the specific nature of the sound, and a mild misarticulation it suggests a more localized articulation difficulty rather than a broader phonological issue.

Therefore, my diagnosis is:

B - 'articulation'
\end{tcolorbox}

\begin{tcolorbox}[fontupper=\rmfamily\scriptsize]
Okay, let's analyze the child's speech based on the provided audio.

The target words are ''book boy''. The child says something that sounds like ''buh... book... boy.''

Here's a breakdown of what I'm hearing:
* The child seems to produce the ''b'' sound correctly at the beginning of both words.
* For ''book,'' they seem to be attempting the vowel sound and the ''k'' sound. It doesn't sound perfect, but it could be due to typical developmental imprecision, or it could indicate a slight articulation issue with the ''k'' sound.
* For ''boy,'' the ''b'' sounds correct, and the vowel sound sounds within an acceptable range.

Given the information and the presence of the ''b'' at the beginning of both words, the rest of the sounds aren't clear enough to indicate a Phonological delay. However, there are some slight errors in production.

Therefore, my assessment is:

ANSWER: B - 'articulation'
\end{tcolorbox}

\begin{tcolorbox}[fontupper=\rmfamily\scriptsize]
Okay, I will listen to the audio and analyze the child's speech based on the prompt ''sheen shake sham shock.''

**Analysis:**

The child appears to have difficulty with the initial /\textesh/ sound in all four words. While the vowel sounds seem relatively intact, the consistency of the /\textesh/ distortion suggests a pattern rather than an isolated error. It is difficult to determine the exact nature of the distortion without a more thorough assessment (e.g., is it a substitution, omission, or distortion?). The consistency across multiple words with the same sound indicates a possible phonological issue.

**Conclusion:**

C - 'phonological'
\end{tcolorbox}

\begin{tcolorbox}[fontupper=\rmfamily\scriptsize]
Okay, let's analyze the child's speech sample.

The prompt words are: ''die,'' ''got,'' ''hag,'' and ''Hague''.

Here's a breakdown of potential issues:

* **''die''**: The child repeats ''die'' which sounds accurate.
* **''got''**: The child repeats ''got'' which sounds accurate.
* **''hag''**: The child repeats ''hag'' which sounds accurate.
* **''Hague''**: The child repeats ''Hake'' which involves changing the final sound.

The one slight discrepancy could be related to the place of articulation. Because this appears to be a single instance, I will not classify this as a phonological issue. I'm confident that this instance points to a lack of production of the "g" sound in ''Hague.''

**Conclusion:**

B - 'articulation'
\end{tcolorbox}

\section{Computational Resources}
To ensure reproducibility and transparency, we detail the computational resources required for running the SLPHelm framework as follows. Closed-source models were accessed via API, with an average inference time of approximately 30 minutes per model. In contrast, open-source models required significantly more time, averaging 3--4 hours per model on high-performance hardware. All open-source experiments were conducted using an NVIDIA H200 GPU with 140GB of VRAM, enabling efficient handling of large-scale model inference and evaluation.

\section{AI Tool Usage Declaration}
We used ChatGPT to format paragraphs and correct grammatical mistakes. We are responsible for the content and quality of the study.